\documentclass[lettersize,journal]{IEEEtran}
\usepackage{amsmath,amsfonts}
\usepackage[ruled,linesnumbered,vlined]{algorithm2e}  
\usepackage{array}
\usepackage[caption=false,font=footnotesize,labelfont=sf,textfont=sf]{subfig}
\usepackage{textcomp}
\usepackage{stfloats}
\usepackage{url}
\usepackage{verbatim}
\usepackage{graphicx}
\usepackage{cite}

\usepackage{booktabs}
\usepackage{makecell}
\usepackage{colortbl}
\usepackage{xcolor}
\usepackage[normalem]{ulem}

\newcommand{\del}[1]{}
\newcommand{\rev}[1]{#1}

\begin{document}

\title{Knowledge-Guided Memetic Algorithm for Capacitated Arc Routing Problems with Time-Dependent Service Costs}


\author{Qingya Li, Shengcai Liu,  ~\IEEEmembership{Member,~IEEE,} Wenjie Chen, Juan Zou, Ke Tang,  ~\IEEEmembership{Fellow,~IEEE,} Xin Yao, ~\IEEEmembership{Fellow,~IEEE}
\thanks{This work was supported in part by the Guangdong Major Project of Basic and Applied Basic Research under Grant 2023B0303000010, in part by the National Natural Science Foundation of China (NSFC) under Grants 62502192, 62250710682 and 72401105, and in part by an internal grant of Lingnan University. The authors are grateful to Prof. Pietro Simone Oliveto for his valuable comments and constructive suggestions, which helped improve the quality of this paper.  (\textit{Corresponding author: Shengcai Liu.})}
\thanks{Qingya Li, Shengcai Liu and Ke Tang are with the Guangdong Key Laboratory of Brain-Inspired Intelligent Computation, Department of Computer Science and Engineering, Southern University of Science and Technology, Shenzhen 518055, China (e-mail: liqy2020@mail.sustech.edu.cn, liusc3@sustech.edu.cn, tangk3@sustech.edu.cn).}
\thanks{Wenjie Chen is with  the School of Information Management, Central China Normal University, Wuhan, China (e-mail: chenwj6@ccnu.edu.cn)}
\thanks{Juan Zou is with the Hunan Key Laboratory of Computational Intelligence, Xiangtan University, Xiangtan, China (e-mail: zoujuan@xtu.edu.cn).}
\thanks{Xin Yao is with the School of Data Science, Lingnan University, Hong Kong SAR, China (e-mail: xinyao@ln.edu.hk).}
\thanks{\copyright\ 2026 IEEE. Personal use of this material is permitted. Permission from IEEE must be obtained for all other uses, in any current or future media, including reprinting/republishing this material for advertising or promotional purposes, creating new collective works, for resale or redistribution to servers or lists, or reuse of any copyrighted component of this work in other works.}

}

\markboth{Author Accepted Manuscript: IEEE Transactions on Emerging Topics in Computational Intelligence}%
{Li \MakeLowercase{\textit{et al.}}: Knowledge-Guided Memetic Algorithm for CARPTDSC}


\maketitle

\begin{abstract}

The capacitated arc routing problem with time-dependent service costs (CARPTDSC) is a challenging  combinatorial optimization problem that arises from real-world applications. CARPTDSC has two main challenges  in reducing computational time.
First,  it is an NP-hard problem.
 Second, the time-dependent service costs of tasks require frequent evaluations during the search process, significantly increasing   the computational burden. These challenges make it difficult  for existing algorithms to perform  efficient searches, often resulting in limited efficiency.  
 To address these issues, this paper proposes a  knowledge-guided memetic algorithm with  golden section search and negatively correlated search (KGMA-GN), where two knowledge-guided strategies are introduced to improve efficiency. First,  a knowledge-guided  initialization strategy (KGIS) is proposed to generate high-quality initial solutions to speed up convergence. Second, a knowledge-guided small-step-size local search strategy (KGSLSS) is proposed to filter out invalid moves, thereby reducing unnecessary evaluations.
 Experimental results on five benchmark test sets, including both small- and larger-scale instances, demonstrate  that KGMA-GN achieves higher search efficiency than the state-of-the-art methods. Moreover, the ablation study further confirms that the knowledge-guided local search operators in KGSLSS can significantly reduce runtime compared to traditional operators, especially for the knowledge-guided swap operator, which achieves more than a tenfold improvement in speed.
\end{abstract}

\bstctlcite{IEEEexample:BSTcontrol} 
\section{Introduction} \label{sec:intr}
\IEEEPARstart{O}{ver} the past few decades, the capacitated arc routing problem
(CARP) \cite{golden1981capacitated,jia2025neural} has drawn considerable attention within the computational  intelligence community \cite{zhang2023rocash2, tang2009memetic}, owing to its extensive real-world applications,    
 such as garbage collection \cite{jin2020planning} and 
street sweeping \cite{bodin1978computer}.
While the standard CARP  assumes a fixed service cost, many real-world applications involve costs that vary over time.
To better model such scenarios, the CARP with time-dependent service costs (CARPTDSC) has been proposed,
 which was originally inspired by the winter gritting applications~\cite{tagmouti2007arc,tagmouti2010variable}.

\begin{figure}[htbp]
    
		\centering
        \footnotesize
		\subfloat[  Three-segment linear function]{
			\begin{minipage}[t]{0.48\linewidth}
				\hspace*{-0.5cm}
				\includegraphics[width=2in]{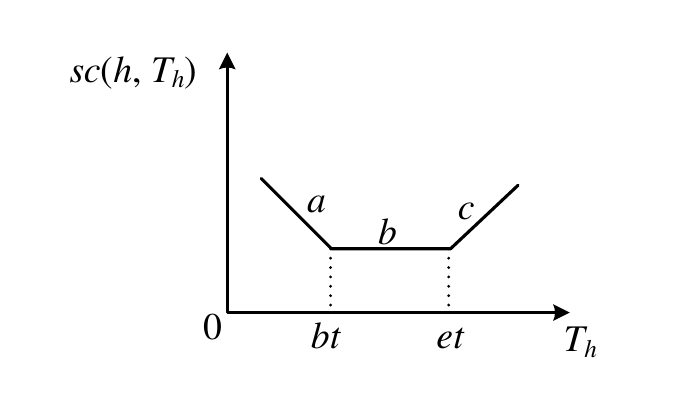}
				\label{fig:3seg}
			\end{minipage}%
		}%
		\subfloat[  Two-segment linear function]{
			\begin{minipage}[t]{0.48\linewidth}
				\hspace*{-0.5cm}
				\includegraphics[width=2in]{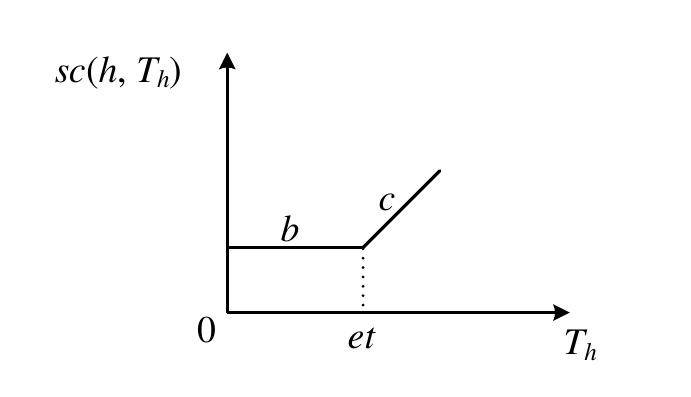}
				 \label{fig:2seg}
			\end{minipage}%
		}%
		
	\centering
	\caption{Schematic diagram of the three-segment and two-segment linear functions between the service cost of task $ h $ and its service beginning time $T_h$. \(a\), \(b\), and \(c\)  represent the decreasing,  unchanged, and increasing segments, respectively.}
	\label{fig0}
\end{figure}

Specifically, the winter gritting application typically takes place on an urban road network, which is modeled  as a connected directed graph. A fleet of homogeneous vehicles, each with a capacity $Q$ of deicing material (e.g., salt), departs from a central depot at different times to clear key roads (i.e., required arcs or tasks). Each task has a specific demand and incurs a service cost (e.g., time or material consumption). Other roads (i.e., travel arcs) are used only for passing and incur  travel costs (e.g., time consumption). The service cost of each task is time-dependent, varying with its  service beginning time due to factors like snow accumulation and traffic conditions. It is always modeled by a three- or two-segment linear function  \cite{tagmouti2007arc, tagmouti2010variable, tagmouti2011dynamic}, as shown in Figure \ref{fig0}.  The objective of CARPTDSC is to determine  vehicle routes and departure times that minimize the total cost.

		
		

CARPTDSC is a complex planning problem with two  major computational challenges.  First, it is NP-hard because it can be reduced to the classical CARP, which has been proven to be NP-hard  \cite{golden1981capacitated}. 
This means that the time required to exactly solve a CARPTDSC instance grows exponentially with the instance size.
Second, frequent evaluations are needed during the search process due to the time-dependent nature of service costs, leading to a high computational burden.

These two challenges make it difficult for algorithms to find satisfactory solutions to CARPTDSC within a limited time in practice. Existing algorithms do not adequately address this  challenge.  A key reason is that they search blindly by evaluating a large number of candidate moves without exploiting problem-specific knowledge to identify promising moves. Consequently, many unnecessary route evaluations are performed, and computationally expensive move assessments are repeatedly conducted for low-quality neighboring solutions. This leads to limited efficiency of existing algorithms.
Therefore, improving efficiency for CARPTDSC remains a key concern. Previous studies \cite{arnold2019knowledge, zhang2024knowledge, beullens2003guided} have shown that incorporating problem-specific knowledge  can improve  efficiency in related problems, such as vehicle routing problems (VRPs) \cite{yue2023graph, zheng2024hybrid, fu2025multi, zhang2024peac, huang2025multimodal,wang2023ant}. Motivated by these works, we attempt to embed  problem-specific knowledge of CARPTDSC into the algorithm to guide the search and improve  efficiency.


Memetic algorithms have been widely applied to solve CARP \cite{tang2009memetic} and  CARPTDSC \cite{li2026novel}. In these algorithms, both the initialization strategy and the  local search  play key roles in determining  efficiency. A good initialization strategy can  generate a set of high-quality initial solutions, 
allowing the algorithm to  start  the search from a higher baseline.
The local search  is the main driver of solution quality improvement, but it typically incurs high computational cost due to its high time complexity. Therefore,  reducing the computational time of local search is crucial for improving  efficiency. 

This paper  proposes a knowledge-guided memetic algorithm with golden section search and negatively correlated search (KGMA-GN)  for CARPTDSC. KGMA-GN incorporates problem-specific knowledge into initialization and local search to improve search efficiency.  Specifically,  a knowledge-guided initialization strategy (KGIS) and a knowledge-guided small-step-size local search strategy (KGSLSS) are introduced. KGIS provides a high-quality initial population, and KGSLSS prunes ineffective moves, avoiding unnecessary evaluations. The main contributions are summarized as follows.

 \begin{itemize}
\item A significant limitation of existing algorithms for CARPTDSC \del{has been identified, namely} \rev{is their limited} search efficiency. This aspect, to the best of our knowledge, remains largely unaddressed in current research. Experimental results further corroborate the substantial search efficiency bottleneck present in existing algorithms. 

\item  To efficiently solve CARPTDSC, KGMA-GN is introduced with two novel components: KGIS and  KGSLSS. KGIS helps the algorithm start with high-quality initial solutions, and 
KGSLSS guides the search toward more promising regions by filtering out failed moves. 
 
\item Experiments on five test sets, including both small- and larger-scale instances,  show  that KGMA-GN indeed improves  efficiency. Specifically, KGMA-GN performs better than other stochastic algorithms in average solution quality and runs faster on all test sets. 
Moreover, the ablation study shows that KGSLSS can greatly reduce runtime, especially for the knowledge-guided swap operator. In particular, this operator is over ten times faster than the traditional swap operator on all test sets.
 \end{itemize}

The remainder  of this paper is organized as follows. Section \ref{sec:rw}    reviews   the  related work. Sections \ref{sec:def} and \ref{sec:meme} introduce the CARPTDSC problem and the proposed algorithm KGMA-GN, respectively. Experimental studies are presented in Section \ref{sec:expe} followed by the conclusion and future work in Section \ref{sec:conc}.

\section{Related work}
\label{sec:rw}
Section \ref{rwc} presents related work on CARPTDSC. The research on memetic methods and knowledge-guided algorithms is introduced in Section \ref{rwa}.

\subsection{Related Work on CARPTDSC}
 \label{rwc}
  
There is limited research  on CARPTDSC.  Tagmouti et al. \cite{tagmouti2007arc} first modeled the winter gritting applications as CARPTDSC where the relationship between the service cost of each task and its service beginning time was modeled by a three-segment linear function. Therefore, this kind of CARPTDSC is called the three-segment linear function problem (3LP) \cite{li2026novel}, which was solved by an exact algorithm called column generation (CG). However, CG   could only handle small-scale instances with up to 40 tasks. Later, Tagmouti et al. \cite{tagmouti2010variable} simplified the relationship to a two-segment linear function by setting the optimal vehicle departure time to zero, resulting in the two-segment linear function problem (2LP) \cite{li2026novel}. They proposed a variable neighborhood descent (VND) algorithm to solve larger-scale CARPTDSC instances. 
  In a follow-up study, Tagmouti et al. \cite{tagmouti2011dynamic} investigated  a dynamic variant  of CARPTDSC, where the optimal service time interval changes according to weather report updates. VND was adapted to solve this problem. 
 To better reflect reality, Li et al. \cite{li2026novel} extended CARPTDSC by assuming that service time was time-dependent and proposed a  effective memetic algorithm called MAENS-GN. Other related problems like the time-dependent CARP \cite{jin2020planning,ahabchane2020mixed,vidal2021arc} are outside the scope of our study, so we will not discuss them in detail here.

 All the aforementioned algorithms for CARPTDSC suffer from limited efficiency.  
 A key reason is that their search is blind, which  often results in ineffective search. To address this limitation, we propose a more efficient algorithm KGMA-GN to guide search and reduce ineffective search operations.
 
  
\begin{algorithm}[!htbp]
	\caption{The General Framework of the Memetic Algorithm.} 
	\label{alg:meme}
	Generate an initial population using the initialization strategy\;
	\While{the stopping condition is not satisfied}
	{
		Generate a new population using the crossover operator\;
		\ForEach {individual $ \in $ the new population}
		{
			Perform a local search strategy on $individual$ with a predefined probability.
		}
		
	}
\end{algorithm}


\subsection{Related Work on Memetic and Knowledge-Guided Algorithms}
 \label{rwa}
The memetic algorithm \cite{qi2015decomposition, dong2022cell,boutaib2025cross} is a commonly used method for solving routing problems \cite{beke2023routing, wen2024transfer, liu2023good}. It combines genetic algorithms with local search, as shown in Algorithm \ref{alg:meme}. Here, we review the state-of-the-art memetic algorithms for CARP and their main strategies for improving search effectiveness or efficiency. 
Tang et al. \cite{tang2009memetic} proposed a large-step-size local search operator  to effectively explore the neighborhood.
Shang et al. \cite{shang2018memetic} introduced   an extension step search to explore promising solutions nearby and a statistical filter   to filter out low-quality solutions in large-scale CARP. 
Li et al. \cite{2021Memetic} designed  a non-smooth penalty to help generate better  solutions.
Vidal et al. \cite{vidal2021arc} adopted efficient data structures for fast travel time queries, and used lower bounds to filter unpromising moves 
 in the time-dependent CARP.

Recent studies have further investigated how knowledge can be 
incorporated into evolutionary and memetic learning frameworks 
to improve adaptation and search efficiency. Hou et al. \cite{hou2017evolutionary}
proposed an evolutionary transfer reinforcement learning framework 
for multiagent systems, where knowledge transfer between related 
agents and tasks was exploited to accelerate learning and improve 
policy adaptation. Hou et al. \cite{hou2021advances} investigated memetic automata 
and discussed their potential for developing autonomous agents 
with adaptive learning capabilities. These studies highlight the 
importance of knowledge reuse and evolution in enhancing the 
efficiency of intelligent search and learning processes.

Problem-specific knowledge can be used to guide the search  and improve efficiency \cite{zhang2024knowledge}. 
Beullens et al. \cite{beullens2003guided}  adopted neighbor list knowledge to reduce the search space in CARP.
Arnold et al. \cite{arnold2019knowledge} introduced problem-specific knowledge, such as the width and depth of routes,  to quickly assess solution quality in VRPs.
Lan et al. \cite{lan2021region}  used distance-based knowledge to define promising regions in large-scale waste collection problems.
Zhang et al. \cite{zhang2024knowledge} applied  node distribution knowledge  to adaptively select  insertion rules in VRPs. 
Meng et al. \cite{meng2026uniteformer} proposed a Transformer-based neural solver that jointly models node and 
edge knowledge to capture richer structural features of routing 
problems.

The memetic algorithm is a general framework for solving routing problems, but ineffective search often arises due to its blind nature. This work introduces a type of knowledge that has not been explored in existing studies, specifically tailored for CARPTDSC. We aim to incorporate this knowledge into the memetic algorithm
to improve efficiency.

\section{The definition of CARPTDSC}
\label{sec:def}
CARPTDSC is defined on a directed graph $ G=(V, A) $, where $ V $ denotes the  set of vertices  and $ A $ is the  set of arcs.  $ A $ can be divided into  two subsets: the set of required arcs  $ A_r $ and  the set of travel arcs $ A_d $. Each required arc $  e \in A_r $, also referred to as a task, is usually   represented by a positive integer and must be serviced by a vehicle. The required arc $ e $ is associated with eight attributes: $ tail(e) $, $ head(e) $, $ d(e) $, $ len(e) $, $ dt(e) $, $ st(e) $, $ dc(e) $, and $ sc(e, T_e) $ which represent the tail vertex, head vertex, demand, length, travel time, service time, travel cost, and time-dependent service cost, respectively. Here, $ sc(e, T_e) $ varies with the service beginning time  on arc $ e $, i.e., $ T_e $. Each travel arc $ e \in A_d $ does not require service, and   usually has five attributes: $ tail(e) $, $ head(e) $, $ len(e) $, $ dt(e) $, and $ dc(e) $ which denotes the tail vertex, head vertex, length, travel time, and travel cost, respectively. The depot is a special vertex in $ V $, labeled by 0, and is also called the dummy task. The notations in the problem definition are summarized in Table \ref{tab:v}.

\begin{table}[!htb]	
	\scriptsize
	\caption{Notations in the problem definition}
	\label{tab:v}	
	\centering
	\noindent \makebox[\textwidth][l]
	{
		\begin{tabular}{lp{6.5cm}}
			\hline  \noalign{\smallskip} 
			$G$ & Directed graph        \\
                $V$ & Set of vertices                      \\
                $A $ & Set of arcs                      \\
                $A_{r} $ & Set of the required arcs                      \\
                $A_{d} $ & Set of the travel arcs                      \\
                $tail(e)$ &Tail vertex of the arc \(e\)     \\
                $head(e)$ & Head vertex of the arc \(e\)     \\
                $d(e) $ & Demand of the required arc \(e\)                      \\
                $len(e) $ &  Length of the arc \(e \)                    \\
                $dt(e) $ & Travel time of the arc \(e\)                      \\
                $st(e) $ &   Service time of the required arc \(e\)              \\
                $dc(e) $ &    Travel cost of the arc \(e \)                    \\
                $T_{e} $ &   Service beginning  time  on the required arc \(e\)                    \\
                $sc(e, T_{e}) $ &  Time-dependent service cost of the required arc \(e\) at  \(T_{e} \)                    \\
                
                $Q $ &   Capacity of the vehicle                    \\         
                $ 0$ &   Depot                    \\
                $R_{k} $ & The \( k\)-th route in the routing plan \(X\)                      \\
                $ t_{k}$ &  Vehicle departure time in \(R_{k} \)                    \\
                $S $ &   Feasible solution                    \\
                $TC(S) $ &   Total cost of the feasible solution \(S\)                    \\
                $ X$ &    Routing plan                   \\
                $ T$ &  Vehicle departure times              \\                
                
                $ m$ &  Number of the routes or vehicles in the routing plan \(X\)                     \\
                $C(R_{k}, t_{k}) $ &  Cost of the route \(R_{k}\) at vehicle departure time \(t_{k}\)                     \\
                                 
                $ e^{k}_{i}$ &    The \(i\)-th  arc in \(R_{k} \)                  \\
                $ l_{k}$ &   Number of the required arcs in \(R_{k}\)                    \\
                $p_{k} $ &  Number of the travel arcs in \(R_{k}\)                     \\                                 
                $|A_{r}| $ &  Number of all required arcs in \(A_r\)                     \\
                $PT $ &   Planning horizon                    \\               
			\hline  \noalign{\smallskip}
		\end{tabular}
	}
\end{table}

\begin{figure}[htbp]
	\centering
	\includegraphics[width=0.35\textwidth]{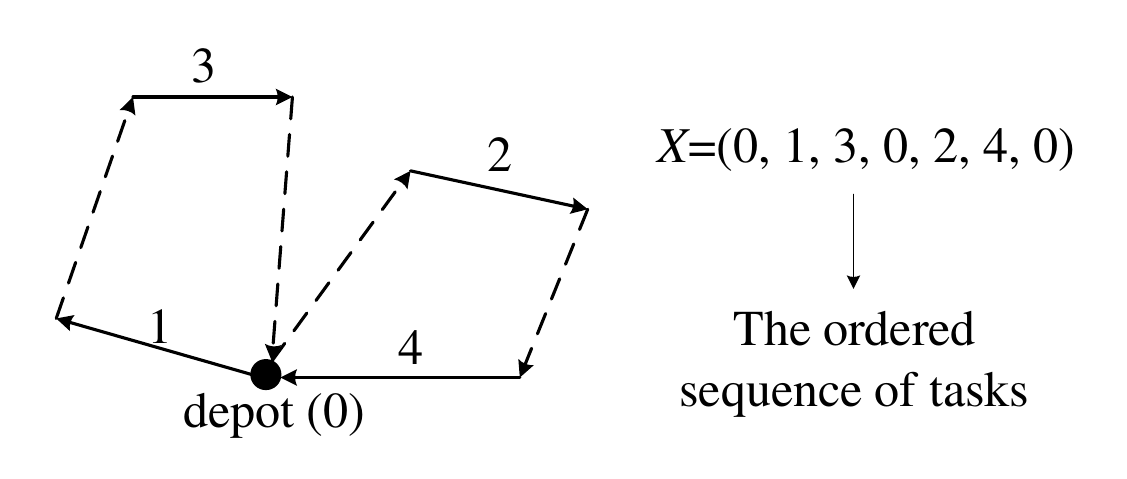}
	\centering
	\caption{An example of the routing plan. The solid  and  dotted lines represent the required arcs (tasks) and the deadheading links, respectively.}
	\label{fig_r}
\end{figure}

Initially, a homogeneous fleet of $ m $ vehicles with capacity $ Q $ is stationed  at the depot. 
Each vehicle $ k $ departs from the depot  at a specified departure time $ t_k $, serves a subset of tasks, and eventually  returns to the depot. The path traversed by  vehicle $ k $  is called a route $ R_k $. Here, $ R_k=(e_0^k, e_1^k, ..., e_{l_k}^k, e_{l_k+1}^k)$, where $ l_k $ is the number of tasks in the route, and $ e_0^k $ and $ e_{l_{k}+1}^k $ represent the sole depot. In addition to the  tasks and the depot,  $ R_k $ also includes a set of travel arcs, denoted as $ (e_{l_{k}+2}^k, \ldots, e_{l_{k}+p_k+1}^k) $, where $ p_k $ is  the number of travel arcs in $ R_k $. These travel arcs are derived  from the shortest paths between adjacent required arcs (i.e., deadheading links), computed using  Dijkstra's algorithm \cite{dijkstra1959note}. The collection of all routes forms a routing plan $ X $. Figure \ref{fig_r} shows an example of the routing plan that $ X=(0, 1, 3, 0, 2, 4, 0)  $ includes two routes: (0, 1, 3, 0) and (0, 2, 4, 0).

The objective of CARPTDSC is to determine the routing plan $ X $ and vehicle departure times $ T $ that minimize the total cost $TC$, i.e., the total time consumption \cite{li2026novel}. The objective function  can be formulated  as:
\begin{equation} \label{equ:1}
min \quad TC(S)=\sum_{k=1}^{m}C(R_k, t_k),
\end{equation}
where $ S $ is a solution defined as $ S=(X, T) $. Specifically,  $ X=(R_1,R_2,\ldots,R_ m) $ and $ T=(t_1, t_2,\ldots,t_m) $.  $ C(R_k, t_k) $ denotes the cost of  route $ R_k $ at vehicle departure time $ t_k $ for $k=1,...,m$. It can be expressed as:
\begin{align} \label{equ:2}
	& C(R_k, t_k)=\sum_{i=0}^{l_k+1} sc(e_i^k, T_{e_i^k})+\sum_{i=l_k+2}^{l_k+p_k+1} dc(e_i^k), \notag \\
	& \hphantom{C(R_k, t_k)=\sum_{i=0}^{l_k+1} sc(e_i^k, T_{e_i^k})+{}}\forall k \in \{1, 2, \ldots, m\},
\end{align}
where the  service beginning time  $ T_{e_i^k} $ can be derived from the vehicle departure time $ t_k $  since there is no waiting time in the route \cite{tagmouti2010variable}.


CARPTDSC is also subject to the following constraints:
\begin{align}
&e_0^k = e_{l_k+1}^k = 0, \quad \forall k \in \{1, 2, \ldots, m\},  \label{equ:3} \\
&e_i^k \neq e_{i'}^{k'}, \quad \forall (k, i) \neq (k',i'), \notag \\
&\hphantom{\neq {}} \forall k,k' \in \{1, 2, \ldots, m\}, \forall i, i' \in \{ 1, \ldots, l_k\}, \label{equ:4} \\
&e_i^k \neq \operatorname{inv}(e_{i'}^{k'}), \quad \forall(k,i)\neq (k',i'), \notag\\
&\hphantom{\neq {}} \forall k,k' \in \{1, 2, \ldots, m\}, \forall i,i' \in \{1, \ldots, l_k\}, \label{equ:5} \\
&\sum_{k=1}^{m} l_k = |A_r|, \label{equ:6} \\
&\sum_{i=1}^{l_k} d(e_i^k) \leq Q, \quad \forall k \in \{1, 2, \ldots, m\}, \label{equ:7} \\
&0 \leq T_{e_i^k} \leq PT, \notag\\
&\hphantom{\leq {}} \forall k \in \{1, 2, \ldots, m\}, \forall i \in \{0, 1, \ldots, l_k, l_k+1\}. \label{equ:8}
\end{align}

Constraint (\ref{equ:3}) ensures   that  each vehicle must start and end at the depot. Constraints (\ref{equ:4})-(\ref{equ:6}) specify   that each task must be served exactly once. In constraint (\ref{equ:5}), inv($e$) denotes the inverse direction of arc $e$, if it exists. In constraint (\ref{equ:6}), $|A_r|$ represents the number of  required arcs in the  set $A_r$.
   Constraint (\ref{equ:7}) imposes the capacity constraint, stating that  the total demand of each route must not exceed the capacity of the  vehicle. Constraint (\ref{equ:8}) ensures  that the  service beginning time on each task  falls within the planning horizon [0, $ PT $].

 \begin{algorithm*}[htb]  
 	\caption{Knowledge-Guided Individual Initialization Strategy}  
 	\label{alg:1}  
 	\KwIn{the set of all tasks, $Tasks_{all}$; the size of $Tasks_{all}$, $N$; the absolute value of the slope $ k $, $|k|$;}  
 	\KwOut{an initial routing plan, $X$;}  
 	
 	$X, R \leftarrow \emptyset$; $numInsertedTasks, load \leftarrow 0$; Add depot 0 to  $R$\;
 	
 	\While{$numInsertedTasks \leq N$}
 	{		
 		The current task $task_{cur}\leftarrow$ the last task in $R$\;
		The candidate task set $Tasks_{candi}  \leftarrow \emptyset$\;
		\ForEach{$ task_{a} \in Tasks_{all}$}
		{
           Add $task_{a}$ to  $Tasks_{candi}$ if inserting $task_{a}$ at the end of $R$ does not violate the capacity constraint and time constraint\;
		}		
				
		\tcc{Select the $``nearest”$ task to be inserted into $R$}
			\If{$Tasks_{candi} \neq \emptyset$}
			{
 		$cost_{min}\leftarrow +\infty$;
 		$Tasks_{nearest}\leftarrow \emptyset$;
 		$ Seq_{tmp} \leftarrow R $\;
 		\ForEach{$task_c \in Tasks_{candi}$}
 		{       
 			Add $task_c$ to the end of	$ Seq_{tmp}$\;
 			Calculate the service beginning time on $task_c$ (denoted as $T_{task_c}$) in $ Seq_{tmp} $, and  $t_{gap}(task_c, T_{task_c}) $\;
 			\If{$sp(head(task_{cur}), tail(task_c))+ t_{gap}(task_c, T_{task_c}) \cdot |k| <cost_{min}$}
 			{
 				$cost_{min}\leftarrow sp(head(task_{cur}), tail(task_c))+ t_{gap}(task_c, T_{task_c}) \cdot |k|$\;
 				$Tasks_{nearest} \leftarrow \{task_c\}$\;				
 			}
 			\ElseIf{$sp(head(task_{cur}), tail(task_c))+ t_{gap}(task_c, T_{task_c}) \cdot |k|=cost_{min}$}
 			{
 				$Tasks_{nearest}\leftarrow Tasks_{nearest}\cup  \{task_c\} $\;
 				
 			}
 			$ Seq_{tmp} \leftarrow Seq_{tmp} \backslash \{task_c\}$
 		}
 		$task_{next}\leftarrow$  randomly select  one task from $Tasks_{nearest}$ \;
 		Add $task_{next}$ to the end of $R$\;
 		$load \leftarrow load + d(task_{next})$\;
 		$Tasks_{all}\leftarrow Tasks_{all} \backslash \{task_{next}\}$\;
 		$numInsertedTasks \leftarrow numInsertedTasks+1$\;
		}
		\Else
 		{           
        \tcc{Create a new route} 
        
          \If{$numInsertedTasks\neq N$}
			{           
            Add $ R $ to the end of $ X $\;	
 			Reset $ R \leftarrow \emptyset$,  and then add depot 0 to $ R $\;	
 			$ load\leftarrow 0 $\; 
            } 	
            \tcc{ Terminate the construction }
            \Else
            {
			Add depot 0 to the end of $R$\;
            Add $ R $ to the end of $ X $\;	
            break\; 			 
            }

 		}
 	}
 	
 	\Return $X$ 
 \end{algorithm*}

\section{The proposed  algorithm KGMA-GN}
\label{sec:meme}
 This section first introduces the problem-specific knowledge in Section \ref{sec:knowledge}. The knowledge-guided initialization strategy (KGIS) and knowledge-guided small-step-size local search strategy (KGSLSS) are then described in Sections \ref{sec:init} and  \ref{sec:local}, respectively. Finally, Section \ref{sec:kgma} presents the overall framework of KGMA-GN.

Before presenting KGMA-GN, we briefly recall the main time-related notations used in the section. Specifically, $t_k$ denotes the departure time of vehicle $k$, and $T_e$ denotes the service beginning time of task $e$. $T=(t_1,t_2,\ldots,t_m)$ denotes the vehicle departure times, and $PT$ denotes the planning horizon.

 \begin{figure}[htbp]
	\centering
	\includegraphics[width=0.31\textwidth]{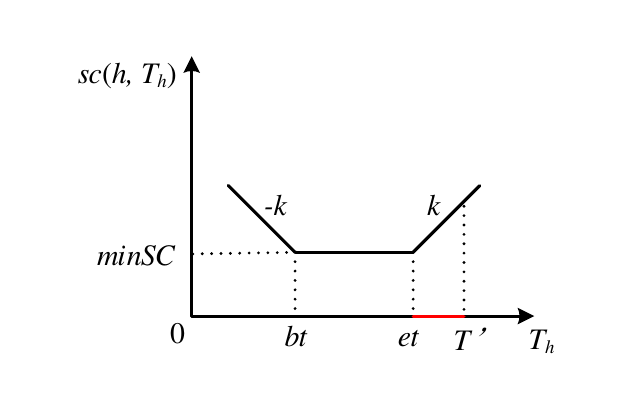}
	\centering
	\caption{An example \del{for} \rev{of} $ t_{gap}(h, T_h) $.  The red line ($ T'-et $) is $ t_{gap}(h, T_h) $.}
	\label{fig_gap}
\end{figure}

\subsection{Problem-Specific Knowledge}
\label{sec:knowledge}
We use the time gap as the problem-specific knowledge.

\textit{Definition 1}: (Time Gap) Let $h$ denote a task and $T_h$ denote its service beginning time. $[bt, et]$ is the optimal service time interval of task $h$, indicating that task $h$ achieves the lowest service cost if it starts being served within this time interval. 
The time gap between $T_h$ and  $[bt, et]$ is defined as $t_{gap}(h,T_h)$. The function $t_{gap}(h,T_h)$ is defined as:
\begin{equation}
\label{equ:10}
t_{gap}(h,T_h)=
\left\{
\begin{aligned}
&|T_h-bt|, \quad T_h <bt \\
&|T_h-et|,  \quad T_h > et \\
&0,  \quad  bt\leq  T_h \leq et.
\end{aligned}
\right.
\end{equation}

A three-segment linear function is used as an example  in  Figure  \ref{fig_gap}.
In this example, $ T' $ is the current   service beginning time on task $ h $. [$ bt $, $ et $] is the optimal service time interval for task $ h $. $k$ and $-k$ are the slopes, respectively, and $minSC$ denotes the minimum service cost. Moreover, $T_h=T'$ and $T'>et$. Thus,  $ t_{gap}(h,T_h)=T' - et $. The service cost of task $ h $ is given by: $ sc(h, T_h) = minSC + t_{gap}(h, T_h) \cdot |k| $. 
It is observed that a smaller time gap results in a lower service cost and if a solution contains tasks with large time gaps, its overall quality tends to be poor.


\subsection{The Knowledge-Guided Initialization Strategy}
\label{sec:init}
KGIS adopts  the initialization framework from \cite{tang2009memetic} to generate an initial population with $psize$ different individuals. Each individual (i.e., the routing plan $X$) is constructed by problem-specific knowledge as shown in Algorithm \ref{alg:1}. This algorithm is a constructive heuristic approach that generates a routing plan $ X $ by iteratively  constructing  route $ R $.  Each route  $R$ is constructed   by iteratively inserting an unserved task into the route, guided by problem-specific knowledge.

Specifically,  $X$, $R$, $numInsertedTasks$, and $load$ are first initialized to \(\emptyset\) or 0, and  depot 0 is added to $R$ (line 1). Here,  $numInsertedTasks$ tracks  the number of tasks  inserted into $X$ or $R$, and $load$ records  the total demand of  route $R$. 
 Then,  $N$ tasks are iteratively inserted into $X$ and $R$ (lines 2-31). The  construction ends  when all tasks have been added to $ X $ (lines 28-31). 

\begin{figure*}[htbp]
		\centering
	\noindent \makebox[\textwidth][l]{
		\subfloat[A single insertion move]{
			\begin{minipage}[t]{0.335\linewidth}
							\centering
				\includegraphics[width=2.1in]{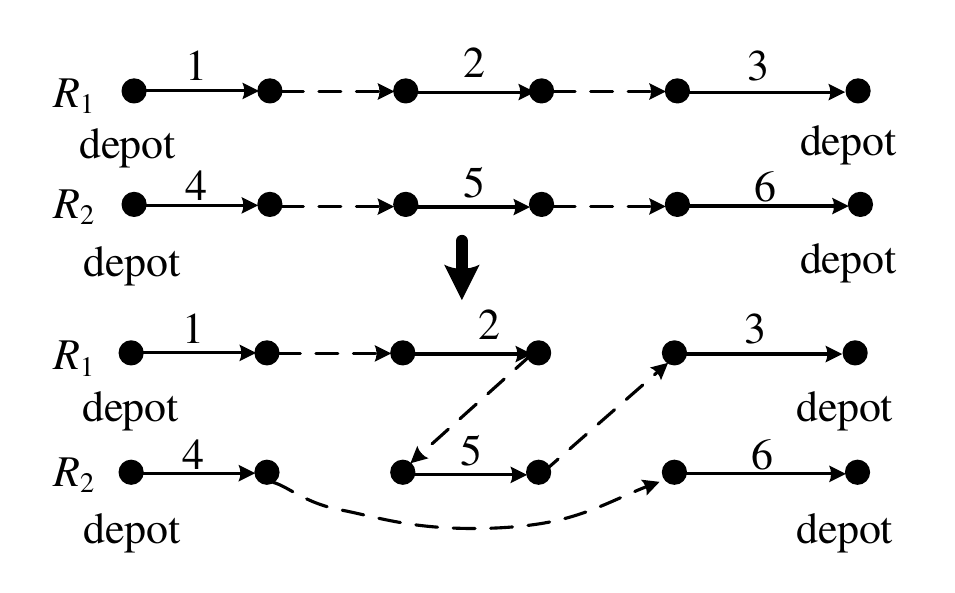}
				
			\end{minipage}%
		}%
		\subfloat[ A double insertion move]{
			\begin{minipage}[t]{0.335\linewidth}
				\centering
				\includegraphics[width=2.1in]{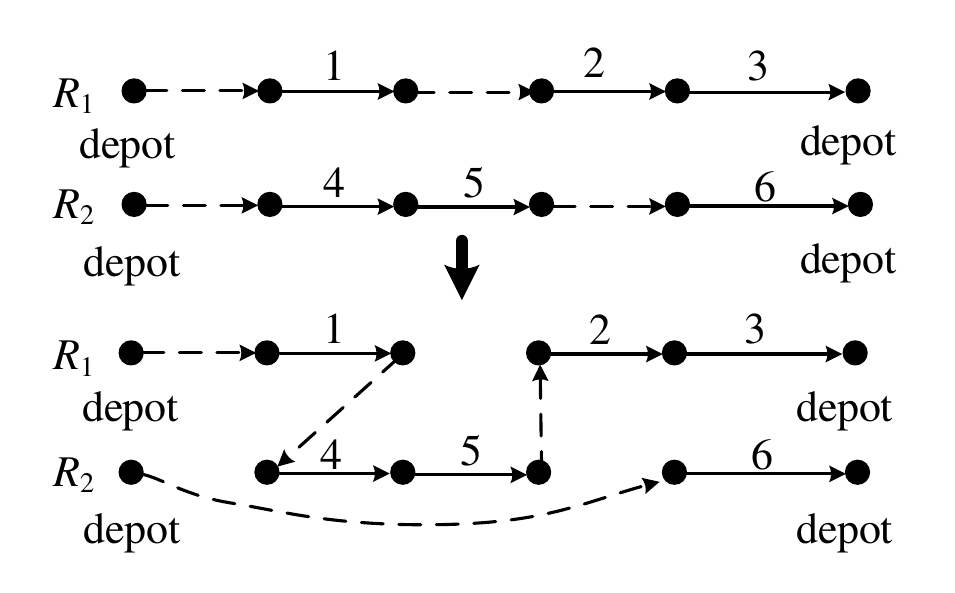}
				
			\end{minipage}%
		}%
	\subfloat[ A swap move]{
			\begin{minipage}[t]{0.335\linewidth}
				\centering
				\includegraphics[width=2.1in]{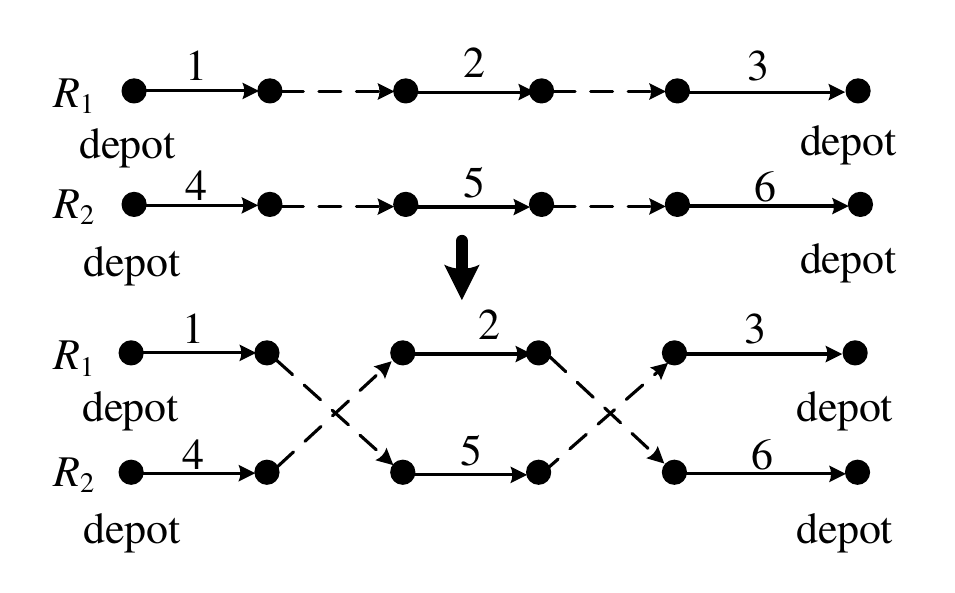}
				
			\end{minipage}%
		}%

	}
	\centering
	\caption{Examples of the three moves. Solid lines represent required arcs (tasks), and dotted lines represent deadheading links. All depot symbols in each move represent the same depot. A move is successful if it leads to a better routing plan; otherwise, it is failed.}
	\label{fig:move}
\end{figure*}

In each iteration of task insertion, the last task in $R$ is first set as the current task $task_{cur}$ (line 3). Then, the candidate task set $Tasks_{candi}$ for the next insertion into $R$ is calculated (lines 4-6). 
After calculating $Tasks_{candi}$, the next step
depends on whether $Tasks_{candi}$ is empty.  If $Tasks_{candi}$ is not empty, a task “nearest" to  $task_{cur}$ is selected and inserted at the end of $R$ (lines 7-22). Specifically, $cost_{min}$, $Tasks_{nearest}$, and $Seq_{tmp}$ are first initialized (line 8). Here, $cost_{min}$ represents the minimum “distance” to $task_{cur}$. $Tasks_{nearest}$ stores  tasks with the minimum “distance” to $task_{cur}$, and $Seq_{tmp}$ is a temporary sequence. Then, $Tasks_{nearest}$ is computed (lines 9-17). Specifically, for each task $task_c$ in  $Tasks_{candi}$ (line 9), it is added to the end of $Seq_{tmp}$ (line 10). Then, the service beginning time on $task_c$ (denoted as $T_{task_c}$) is calculated, along with $t_{gap}(task_c, T_{task_c})$ (line 11). Next, the cost (i.e., “distance”) is computed as the greedy criterion: $sp(head(task_{cur}), tail(task_c))+ t_{gap}(task_c, T_{task_c}) \cdot |k|$, where  $sp(head(task_{cur}), tail(task_c))$ is the shortest travel distance from the head vertex of $task_{cur}$ to the tail vertex of $task_c$. 
If the cost is smaller than $cost_{min}$, both $cost_{min}$ and $Tasks_{nearest}$ are updated (lines 12-14). If the cost equals $cost_{min}$, $task_c$ is added to $Tasks_{nearest}$ (lines 15-16). Then, $Seq_{tmp}$ is updated (line 17). After calculating $Tasks_{nearest}$, one task is randomly selected from it and inserted at the end of $R$ (lines 18-19).  Then, $load$, $Tasks_{all}$, and $numInsertedTasks$ are  updated (lines 20-22). 
If $Tasks_{candi}$ is empty,  no feasible task can be added to $R$. In this case,  it is checked whether all tasks have been inserted into either $X$ or the current route $R$. 
if not all tasks have been inserted, a new route $R$ is created (lines 24--27). Specifically, the current route $R$ is first appended to the end of $X$ (line 25). Then, $R$ is reset to an empty route, depot 0 is inserted into $R$ (line 26), and  \(load\) is updated (line 27). Otherwise, the construction process terminates (lines 28--31). Specifically, depot 0 is appended to the end of $R$ (line 29), the final route $R$ is added to the end of $X$ (line 30), and the construction loop is terminated via the \texttt{\textit{break}} statement (line 31).

 \begin{table*}[]
	\scriptsize
	\caption{Illustration of an invalid search phenomenon by the swap move.}
	\label{tab:expl}
	\centering
	\begin{tabular}{@{}ccccccccc@{}}
		\toprule
		& \multicolumn{4}{c}{Before swapping($ r_i $, $ s_j $)}                               & \multicolumn{4}{c}{After swapping($ r_i $, $ s_j $)}                                          \\ 
		\cmidrule(lr){2-5} \cmidrule(lr){6-9} 
		\multicolumn{1}{l}{} & $ T_h $                & {[}$ bt $, $ et ${]}  & $ t_{gap}(h, T_h) $ & $ sc(h, T_h)$ & $ T_h $   & {[}$ bt $, $ et ${]}   & $ t_{gap}(h, T_h) $ & $ sc(h, T_h) $ \\
		\midrule
		$ r_i $                   & {2}   & {[}1,3{]}     & 0             & 1         & {502} & {[}1,3{]}     & 499           & 500       \\
		$ s_j $                   & {502} & {[}501,503{]} & 0             & 1         & {2}   & {[}501,503{]} & 499           & 500       \\ \bottomrule
	\end{tabular}
\end{table*}

\subsection{Knowledge-Guided  Local Search Strategy}
\label{sec:local}
Traditional  local search operators are first introduced in Section \ref{sec:trals}. Section \ref{sec:kglss} describes KGSLSS in detail.

\subsubsection{Traditional  Local Search Operators}
\label{sec:trals}

Traditional operators are introduced in Section \ref{sec:tlso}. Section \ref{sec:tcto} presents the time complexity of traditional  operators.
Section \ref{sec:phen} describes an invalid search phenomenon.

\paragraph{Traditional Small-Step-Size Local Search Operators} 
\label{sec:tlso}  
\indent Local search is typically  performed   by applying move operators to a candidate routing plan.  These operators usually have  small search step sizes.  Common small-step-size local search operators include single insertion, double insertion, and swap \cite{tang2009memetic}. 


\begin{itemize}   

	\item \textbf{Single insertion move:}  A single task is removed from its current position  and reinserted into  another position of the routing plan or into a new empty route. As  shown in Figure \ref{fig:move}(a), task 5 is removed from route $ R_2 $ and then reinserted into $ R_1 $ after task 2.
	\item \textbf{Double insertion move:}  Two consecutive tasks are removed and then reinserted into another position of the routing plan or into a new empty route. As  shown in Figure \ref{fig:move}(b),   tasks 4 and 5  are removed from route $ R_2 $, and reinserted into $ R_1 $ after task 1.
	\item \textbf{Swap move:}  The positions of  two tasks are exchanged. As  shown in Figure \ref{fig:move}(c), the positions of task 2 in route $ R_1 $ and task 5 in route $ R_2 $ are exchanged.
\end{itemize}


\paragraph{Time Complexity of Traditional  Operators} 
\label{sec:tcto}  

In CARPTDSC,  a move typically affects  the service costs of many tasks in the involved   routes  of the current candidate routing plan. As a result, identifying whether a move is successful  often requires evaluating the involved routes. If the involved routes contain  $ h $ tasks, the time complexity of a single  evaluation is generally   O($ h $).  Given that the neighborhood size is $ \Theta(N^2) $ \cite{tang2009memetic}, where $ N $ is the number of tasks in the instance,   a full neighborhood search for each traditional local search operator has a time complexity of O$(hN^2)$. Therefore,  local search in CARPTDSC is computationally expensive.

\paragraph{An Invalid Search Phenomenon} 
\label{sec:phen} 


Since route evaluation is required to identify failed moves, it leads to unnecessary  time consumption. If failed moves can be identified  in advance, route evaluation can be skipped to save time.  We observe that in traditional local search operators for CARPTDSC, failed moves  often cause significant changes in problem-specific knowledge (i.e., the time gap). This observation will be illustrated with an example in the next paragraph.  As shown in Section \ref{sec:knowledge}, a routing plan with multiple tasks having large time gaps tends to be of low quality. Therefore, if a move significantly increases the time gaps of several tasks, it is likely to be a failed move.

Take the swap move as an example to illustrate the invalid search phenomenon, as shown  in Table \ref{tab:expl}. Let the two involved routes be  $ R_1=(r_1, r_2,\ldots, r_{n_1}) $ and $ R_2=(s_1, s_2,\ldots, s_{n_2}) $. The move swap ($ r_i $, $ s_j $) exchanges  the positions of $  r_i $ and $ s_j $, where $i \in \{1, 2, \ldots, n_1\}  $ and $ j \in \{1, 2, \ldots, n_2\} $. 
The time-dependent cost function of each task $ h \in \{r_i, s_j \}$ is  shown in  Figure \ref{fig_gap}, with $ minSC =1$ and  $ |k|=1 $.  
This move increases the total service cost and   time gap  by 998 from just  these two tasks (i.e., $ r_i $ and $ s_j  $).
Therefore, this move is very likely to fail.


\subsubsection{KGSLSS}
\label{sec:kglss}

 KGSLSS is described  in Algorithm \ref{alg:4}, which  contains three knowledge-guided local search operators: knowledge-guided single insertion (line 1), knowledge-guided double insertion (line 2), and knowledge-guided swap (line 3). Each operator generates its best routing plan, and the overall best one is selected from them (line 4).
 
 \begin{algorithm}[!htb]  
 	\caption{KGSLSS}  
 	\label{alg:4}  
 	\KwIn{a routing plan, $ S $;  the traditional single insertion operator, $si$; the traditional double insertion operator, $di$; the traditional swap operator, $sw$}
 	\KwOut{the locally optimal routing plan, $ S_{best} $}  
 	$ S_1 \leftarrow $  apply Algorithm \ref{alg:3}  to $ S $ and  $si$\;
 	$ S_2 \leftarrow $  apply Algorithm \ref{alg:3}  to $ S $ and $di$\;
 	$ S_3 \leftarrow $  apply Algorithm \ref{alg:3}  to $ S $ and  $sw$\;
 	$ S_{best} \leftarrow $ select the best routing plan from \{$ S_1, S_2, S_3 $\}\;
 	
 	\Return $ S_{best} $\;
 \end{algorithm} 
 
  \begin{algorithm}[!htb]  
 	\caption{Knowledge-Guided Local Search Operator}  
 	\label{alg:3}  
 	\KwIn{a routing plan, $ S $; the traditional local search operator, $ ope $}  
 	\KwOut{the locally optimal routing plan, $ S_{best}$}  
 	$ move_{set} \leftarrow$ get  all possible  moves in the $ ope $\;
 	$ S_{best} \leftarrow S$\;
 	\ForEach{$ move \in move_{set} $}
 	{
 		$ S' \leftarrow $ apply $ move $ to $ S $\;
 		
 		\tcc{Identify the type of move}
 		\If{$move$ satisfies Criterion  1}
 		{
 			The type of  $ move  \leftarrow failed$\;
 		}
 		\Else
 		{
 			\If{move satisfies Criterion  2}
 			{
 				The type of $ move \leftarrow successful$\;
 			}
 			\Else
 			{
 				The type of $ move \leftarrow failed$
 			}
 			
 		}

 		\tcc{Update $ S_{best}$}
 		\If{The type of $ move = successful$}
 		{
 			\If{$ S'$ is better than $ S_{best} $ }
 			{
 				$ S_{best} \leftarrow S'$\;
 			}
 		}
 	}
 	\Return $ S_{best}$\;
 	
 \end{algorithm} 

Each knowledge-guided  operator is transformed from its corresponding traditional operator by Algorithm \ref{alg:3}. 
The \del{move set $move_{set}$ in} \rev{set of moves associated with}   $ope$ is first gotten (line 1).
The locally optimal routing plan $S_{best}$ is then initialized as the current routing plan $S$ (line 2). A neighborhood search  is then performed on  $S$   (lines 3-14). Specifically, for each move in $move_{set}$, it is applied to $S$ to generate a new routing plan $S'$ (line 4). The type of the move  is then identified based on Criterion 1 and Criterion 2 (lines 5-11). If the move is identified as successful, $S_{best}$ is updated with $S'$ (lines 12-14).

\begin{algorithm*}[!htb]  
	\caption{Knowledge-Guided Memetic Algorithm with  Golden Section Search and Negatively Correlated Search}  
	\label{alg:5}  
	\KwIn{an instance, $ inst $;  the size of the population, $ psize $; the number of  generating offsprings per generation, $ osnum $; the number of  generations, $ gnum $; the probability of performing  local search,  $ Pls $ }  
	\KwOut{the best feasible routing plan in the population, $ S_{bf} $; the vehicle departure times,  $T$}	
	\tcc{The first stage:  solving the routing plan }
	Initialize  a population $ pop $ with $ psize $ diverse  individuals by KGIS\;
	Generation counter $ gcounter \leftarrow 0 $\;
	\While{$ gcounter<gnum $}	
	{
		\For{$ i =1 \rightarrow osnum $}
		{
			Randomly select two individuals $ S_1 $, $ S_2 $ as parents from $ pop $\;
			$ S_3 \leftarrow $ apply the crossover operator SBX to  $ S_1 $ and $ S_2 $\;
			Generate a random number $ r $ between 0 and 1\;
			\If{$ r <Pls $}
			{
				$ S' \leftarrow $  apply  KGSLSS (Algorithm \ref{alg:4})   to $ S_3 $\;
				$ S' \leftarrow $  apply large-step-size local search operator Merge-Split to $ S' $\;
				$ S' \leftarrow $  apply KGSLSS (Algorithm \ref{alg:4}) to $ S' $\;
				\If{$ S' \notin pop $}
				{
					$ pop \leftarrow pop \cup S' $\;
				}
				\ElseIf{$ S_3 \notin pop $}
				{	       	
					$ pop \leftarrow pop \cup S_3 $\;
					
				}
			}
			\ElseIf{$ S_3 \notin pop $}
			{
				$ pop \leftarrow pop \cup S_3 $\;
			}	  
		}
		Sort $ pop $ with  stochastic ranking\;
		$ pop \leftarrow $ select the best $ psize $ of individuals from  $ pop $\;	
		$ gcounter \leftarrow gcounter + 1 $\;
	}
	    $ S_{bf} \leftarrow$ the best individual in $ pop $\;	
		\tcc{The second stage: determining the optimal or near-optimal vehicle departure times}
		Get the instance type \(intp\) and the absolute value of the slope $k$ (i.e., $|k|$) from \(inst\)\;	
	
	\If{\(intp = 2LP\)}
	{
		  \textit{m} $\leftarrow$ the number of routes in $S_{bf}$\;
		\For{$i = 1 \rightarrow$  m}
		{
            \(t_i \leftarrow 0\);
		}
        \(T \leftarrow (t_1,t_2,\ldots,t_m)\)\;
	}
	\ElseIf{\(intp = 3LP\) }
	{
		
		\If{$|k| \leq 1$}
		{
			 $ T \leftarrow \boldsymbol{GSS(S_{bf},\ 0,\ PT)}$\;
		}
		\ElseIf{$|k| > 1$}
		{
			 $T \leftarrow \boldsymbol{NCS(S_{bf},\ 0,\ PT)}$\;
		}
	}
	\Return $ S_{bf} $ and  $T$\;
\end{algorithm*}  

\paragraph{Criterion  1}
\label{sec:jc1}  

Section \ref{sec:phen} shows that problem-specific knowledge can be used to identify failed moves. Based on this observation,  we propose Criterion 1 that the total time gap of the most relevant (i.e., the directly affected) tasks  becomes significantly  larger after applying a move. If the move satisfies Criterion 1, it is considered failed.
 Let the set of the most relevant  tasks be $Tasks_{relt} = \{e_1, e_2, \dots, e_p\}$, where $p$ is the number of these tasks in the move.  
For each task $e_i$ in $Tasks_{relt}$,
$t_{gap}(e_i, T_{e_i})$ and $t'_{gap}(e_i, T_{e_i})$ represent its time gap  before and after the move, respectively. Therefore, Criterion 1 can be expressed as: 
\[
\sum_{i=1}^{p} t'_{gap}(e_i, T_{e_i} )-\lambda \cdot \sum_{i=1}^{p} t_{gap}(e_i, T_{e_i} )>0,
 \]
where $\lambda$ is a control parameter. A smaller $\lambda$ makes candidate moves more likely to meet Criterion 1, thereby causing more candidate moves to be classified as failed moves and filtered out, making the filtering rule stricter. Conversely, a larger $\lambda$ makes candidate moves less likely to meet Criterion 1, allowing more candidate moves to proceed to Criterion 2 and making the filtering rule less strict.

The most relevant tasks in Criterion 1 depend on the type of the move. In a single insertion move,  the most relevant task is the one being removed and reinserted (e.g., task 5 in Figure \ref{fig:move}(a)). In a double insertion move, the most relevant tasks are the two consecutive tasks being removed and reinserted (e.g., tasks 4 and 5 in Figure \ref{fig:move}(b)). In a swap move, the most relevant tasks are the two  being exchanged  (e.g., tasks 2 and 5 in Figure \ref{fig:move}(c)).

 \paragraph{Criterion  2}
 \label{sec:jc2}

 Criterion 2 identifies the type of a move by considering only the tasks and deadheading links affected within the involved routes. Specifically, Criterion 2 states that  the difference between the total service cost of the affected tasks after the move and before the move, plus the difference between the total travel cost of the added deadheading links and that of the removed ones, is less than 0. If the move satisfies Criterion 2, the move is considered successful. 
Criterion 2 is expressed as: \[\Delta sc + \Delta dc<0,\] where $ \Delta sc $ is the change in service cost, calculated as the total service cost after the move minus that before the move, considering only  the tasks whose service costs have changed. 
$ \Delta dc $ denotes the change in travel cost resulting from modified deadheading links in the involved routes.

\paragraph{Time Complexity of KGSLSS}
 \label{sec:timecoty}

We first analyze the time complexities of Criteria 1 and 2, and then introduce the time complexity of KGSLSS.

Before performing  local search, the time gap of each task in the routing plan can be precomputed   and stored in an array. 
 Then,  the time gap of each task used in  Criterion  1 to  identify the  type of a move can be directly obtained from the array with constant time complexity, i.e.,  O(1). Therefore, the time complexity of Criterion  1 is O(1).
 
Suppose a move affects the service costs of $k $ tasks in the     involved routes.
 Additionally,  the number of modified deadheading links in  the involved routes is typically fixed and can be treated as a constant. 
 For example, as shown in  Figure \ref{fig:move}(a), the single insertion move  modifies six deadheading links.
 Before the local search begins, the travel cost of the shortest path between each pair of vertices  (i.e., each potential  deadheading link) is  precomputed  by  Dijkstra's algorithm and stored  in an array. 
 In this way,  the travel cost of any deadheading link can be retrieved  in constant time, i.e., O(1).  Therefore, the time complexity of Criterion  2 is O($ k $).

As shown in Algorithm \ref{alg:3}, if the move exhibits the invalid search phenomenon, the time complexity of the evaluation of the involved  routes  is O(1) by Criterion  1. If the move does not exhibit this phenomenon,  the time complexity
is O($ k $) by Criterion  2. Therefore,  the worst-case time complexity of each knowledge-guided local search operator in KGSLSS is O$(kN^2)$. Because $k\leq h$ and the hidden constants in $O(kN^2)$ are no greater than those in $O(hN^2)$ (i.e., the time complexity of  traditional operators), each knowledge-guided local search operator in KGSLSS has lower time complexity than its corresponding traditional local search operator.


\begin{table*}[htbp]
\caption{Positioning of KGMA-GN with respect to MAENS-GN.}
\label{tab:MAENS_KGMA}
\centering
\renewcommand{\arraystretch}{1.15}
\begin{tabular}{lll}
\toprule

\textbf{Component} & \textbf{MAENS-GN} & \textbf{KGMA-GN} \\
\midrule

Solution representation 
& Same 
& Same \\

Route optimization framework 
& MAENS framework 
& MAENS framework \\

Departure-time optimization 
& GSS/NCS 
& GSS/NCS \\

Initialization 
& Traditional greedy initialization
& KGIS \\

Crossover operator 
& SBX 
& SBX \\

Local search 
& Traditional neighborhood search 
& KGSLSS \\

Efficient filtering  rule
& Not available 
& Criterion 1 and Criterion 2 \\

\bottomrule
\end{tabular}
\end{table*}
 \subsection{The Framework of KGMA-GN}
 \label{sec:kgma}

KGMA-GN is developed based on MAENS-GN by incorporating problem-specific knowledge into the memetic search framework. As summarized in Table~III, KGMA-GN inherits several key components from MAENS-GN, including the solution representation, route optimization framework, departure-time optimization strategy, and crossover operator. The main modifications lie in the initialization and local search procedures. Specifically, the traditional greedy initialization is replaced by the proposed knowledge-guided initialization strategy (KGIS), while the traditional neighborhood search is replaced by the knowledge-guided small-step-size local search strategy (KGSLSS). In KGSLSS, two efficient filtering criteria are introduced to avoid unnecessary route evaluations. These knowledge-guided components constitute the main contributions of KGMA-GN and enable a more efficient search than MAENS-GN without changing its overall optimization framework.
 
  KGMA-GN, shown  in Algorithm \ref{alg:5}, is a two-stage algorithm. In the first stage, a knowledge-guided memetic algorithm (KGMA) is used to solve the routing plan (lines 1-21). In the second stage, either golden section search (GSS) \cite{OVERHOLT1967FIBONACCI} \del{and} \rev{or} negatively correlated search (NCS) \cite{tang2016negatively} is applied to determine the optimal or near-optimal vehicle departure times for the routing plan (lines 22-32). 

 In the first stage, 
 an initial population $ pop $ with  $ psize $ diverse individuals is first generated by  KGIS  (line 1). The generation counter $gcounter$ is initialized to 0 (line 2).  The population then evolves  for $ gnum $ generations (lines 3-20). For each generation,  $osnum$ offspring are generated and added  to  $pop$ (lines 4-17).  
Specifically, two individuals are randomly selected from  $ pop  $ as parents (line 5), and an offspring $ S_3 $    is produced by applying   SBX   \cite{tang2009memetic} (line 6). If a random  number $ r $ is less than the threshold  $ Pls $,  local search is performed on $ S_3 $ (lines 7-17). Specifically, the local search  combines KGSLSS with a large-step-size local search operator Merge-Split \cite{tang2009memetic} (lines 9-11). The resulting offspring  is then added to   $ pop $ (lines 12-17). After generating all offspring,  the individuals in $ pop $ are sorted using stochastic ranking \cite{runarsson2000stochastic}, and the best $ psize $ individuals are retained (lines 18-19).  Then,  $gcounter$ is updated (line 20). After all generations, the best individual in the final population, denoted as $S_{bf}$, is selected (line 21). \\
\indent In the second stage, the vehicle departure times $T$ 
are optimized (lines 22-32). The method used to compute  $T$ depends on the instance type, denoted as $intp$. If $ intp $ is 2LP, the relationship between route cost and vehicle departure time is  non-decreasing \cite{li2026novel}, so the optimal  departure time is 0 (lines 23-27). If $ intp $ is 3LP, the choice of method depends on the slope $k$ (lines 28-32). If  $ |k| \leq 1$, the relationship  is (similar) unimodal \cite{li2026novel}, and  GSS \cite{OVERHOLT1967FIBONACCI} is applied to solve  $T$  (lines 29-30). If $ |k|>1 $, the relationship becomes non-unimodal \cite{li2026novel}, and  NCS \cite{tang2016negatively} is employed  to solve  $T$ (lines 31-32).

\section{Experimental studies}
\label{sec:expe}

To evaluate the  efficiency of KGMA-GN, two sets of empirical studies are conducted to compare it with state-of-the-art algorithms. In the first study, the algorithms are compared from the perspective of solution quality  given a predefined time budget. In the second study, the performance of different algorithms is examined in terms of the runtime required to achieve a predefined solution quality. In addition, further empirical analyses are conducted to assess the effectiveness of  two proposed knowledge-guided strategies.

\subsection{Experimental Setup}
\label{sec:setup}
This section  introduces benchmark test sets, compared algorithms, performance metrics, and experimental protocol. 


\begin{table*}[!htb]
	\scriptsize
	\caption{Results on the $2lp$-$gdb$ benchmark test set  in terms of costs of solutions given a predefined time budget.  }
	\label{tab:2}
	\centering
	\noindent \makebox[\textwidth][c]
	{
		\begin{tabular}{@{}ccccccccccc@{}}
			\toprule
			Instance & $|R|$ & LB  & \multicolumn{2}{c}{MAENS-GN} & \multicolumn{2}{c}{KGMA-GN} & \multicolumn{2}{c}{VND} \\ 
			\cmidrule(lr){4-5} \cmidrule(lr){6-7} \cmidrule(lr){8-9}
			&     &                   & Ave(Std)                                     & Best & Ave(Std)                         & Best       & Ave(Std)    & Best  \\
			\midrule
			w-d-l & -   & -   & 3-20-0                                      & -       & -                                  & -    & 23-0-0    & -    \\
			No.best     & -   & -   & -                                            & 22     & -                                  & 23     & -           & 5     \\
			Ave.PDR     & -   & -   & 1.37\%                                      & -      & 1.15\%                             & -           & 8.79\%  & -         \\ 
			\bottomrule
		\end{tabular}
		
	}
\end{table*}

\begin{table*}[!htb]
 	\scriptsize
 	\caption{Results on the $2lp$-$egl$ benchmark test set  in terms of costs of solutions given a predefined time budget.  }
 	\label{tab:4}
 	\centering
 	\noindent \makebox[\textwidth][c]
 	{
 		\begin{tabular}{cccccccccc}
 			\toprule
 			Instance  & $|R|$ & LB    & \multicolumn{2}{c}{MAENS-GN}        & \multicolumn{2}{c}{KGMA-GN} & \multicolumn{2}{c}{VND}  \\
 			\cmidrule(lr){4-5} \cmidrule(lr){6-7} \cmidrule(lr){8-9}
 			&     &                      & Ave(Std)                                  & Best      & Ave(Std)    & Best       & Ave(Std)    & Best  \\
 			\midrule
 			w-d-l  & -   & -     & 19-5-0                                       & -      & -              & -     & 23-1-0         & -      \\
 			No.best      & -   & -     & -                                            & 6      & -               & 23    & -              & 0     \\
 			Ave.PDR      & -   & -     & 4.18\%                     & -      & 3.05\%          & -     & 10.54\%         &-      \\
 			\bottomrule
 		\end{tabular}
 		
 	}
 \end{table*}

\begin{table*}[!htb]
 	\scriptsize
 	\caption{Results on the $3lp$-$gdb$ benchmark test set  in terms of costs of solutions given a predefined time budget. For each instance, the average performance of an algorithm is indicated with a dark background if it is the best among all comparison algorithms. ``$\ddagger$" and ``$\dagger$" indicate KGMA-GN performs significantly better than and equivalently to the corresponding algorithm based on 20 independent runs according to the Wilcoxon rank-sum test with significant level $ p=0.05 $, respectively. }
 	\label{tab:6}
 	\centering
 	\noindent \makebox[\textwidth][c]
 	{
 		\begin{tabular}{@{}ccccccccccccc@{}}
 			\toprule
 			Instance & $|R|$ & LB  & \multicolumn{2}{c}{MAENS-GN}     & \multicolumn{2}{c}{KGMA-GN} & \multicolumn{2}{c}{VND-GN} \\
 			\cmidrule(lr){4-5} \cmidrule(lr){6-7} \cmidrule(lr){8-9}
 			&     &                 & Ave(Std)                                     & Best     & Ave(Std)    & Best    & Ave(Std)   & Best  \\
 			\midrule
 			3lp-gdb1    & 22  & 316 & 618.5(0.90)$\dagger$               & 617.4    & 618.6(0.65)   & 617.6  & 617.8(0.43) \cellcolor{lightgray}  & 617.1 \\
 			3lp-gdb2    & 26  & 339 & 622.4(0.92)$\dagger$ \cellcolor{lightgray}   & 621.2  & 622.3(0.85)\cellcolor{lightgray}   & 621.1  & 630.5(14.07)$\dagger$ \cellcolor{lightgray} & 620.8 \\
 			3lp-gdb3    & 22  & 275 & 289.0(0.00)$\dagger$ \cellcolor{lightgray}   & 289.0   & 289.0(0.00) \cellcolor{lightgray}   & 289.0  & 312.4(5.46)$\ddagger$  & 309.0   \\
 			3lp-gdb4    & 19  & 287 & 745.5(0.88)$\dagger$                         & 744.4  & 745.7(0.74)   & 744.3  & 744.9(0.54)\cellcolor{lightgray}  & 744.2 \\
 			3lp-gdb5    & 26  & 377 & 389.6(9.22)$\dagger$ \cellcolor{lightgray}   & 377.5   & 384.7(4.81)\cellcolor{lightgray}   & 377.5  & 407.8(8.71)$\ddagger$  & 393.0   \\
 			3lp-gdb6    & 22  & 298 & 572.9(14.64)$\dagger$ \cellcolor{lightgray}  & 562.3  & 569.6(12.8)\cellcolor{lightgray}   & 562.4  & 583.5(18.10)$\ddagger$ & 539.3 \\
 			3lp-gdb7    & 22  & 325 & 468.0(3.82)$\dagger$\cellcolor{lightgray}    & 459.4  & 468.9(4.38)\cellcolor{lightgray}   & 459.8  & 493.3(10.21)$\ddagger$ & 469.4 \\
 			3lp-gdb8    & 46  & 348 & 391.7(11.09)$\dagger$ \cellcolor{lightgray}  & 368.8  & 394.0(12.24) \cellcolor{lightgray}  & 372.0  & 405.4(17.83)$\ddagger$ & 383.1 \\
 			3lp-gdb9    & 51  & 303 & 319.8(10.18)$\dagger$ \cellcolor{lightgray}  & 309.0  & 315.6(4.82)\cellcolor{lightgray}   & 309.0  & 353.4(11.63)$\ddagger$ & 329.0   \\
 			3lp-gdb10   & 25  & 275 & 286.6(11.19)$\ddagger$                       & 275.0   & 279.0(8.32) \cellcolor{lightgray}   & 275.0  & 312.5(8.34)$\ddagger$  & 299.0   \\
 			3lp-gdb11   & 45  & 395 & 417.5(5.36)$\ddagger$                        & 411.0  & 409.6(6.07) \cellcolor{lightgray}   & 399.0  & 437.4(9.41)$\ddagger$  & 418.0   \\
 			3lp-gdb12   & 23  & 458 & 479.4(25.84)$\dagger$ \cellcolor{lightgray}  & 458.0  & 487.0(28.5) \cellcolor{lightgray}   & 467.0  & 533.0(17.16)$\ddagger$ & 488.0   \\
 			3lp-gdb13   & 28  & 536 & 615.1(3.39)$\dagger$ \cellcolor{lightgray}   & 609.8  & 616.0(3.95) \cellcolor{lightgray}   & 608.4  & 637.4(12.34)$\ddagger$ & 622.5 \\
 			3lp-gdb14   & 21  & 100 & 167.5(6.13)$\dagger$                         & 154.8   & 165.7(7.92)   & 148.2  & 157.3(9.53)\cellcolor{lightgray}  & 131.2 \\
 			3lp-gdb15   & 21  & 58  & 79.5(1.52)                                   & 76.8   & 80.5 (1.34)   & 78.3   & 78.6(0.64) \cellcolor{lightgray}  & 76.7  \\
 			3lp-gdb16   & 28  & 127 & 137.1(2.61)$\dagger$\cellcolor{lightgray}    & 133.1  & 136.2(2.77)\cellcolor{lightgray}   & 132.8  & 138.2(0.00)$\ddagger$     & 138.2 \\
 			3lp-gdb17   & 28  & 91  & 113.3(1.50)$\dagger$                         & 109.9  & 113.6(1.14)   & 111.3  & 111.7(0.38)\cellcolor{lightgray}  & 111.2 \\
 			3lp-gdb18   & 36  & 164 & 167.6(3.34)$\dagger$\cellcolor{lightgray}    & 164.0  & 166.4(2.15)\cellcolor{lightgray}   & 164.0  & 178.4(6.74)$\ddagger$  & 168.0   \\
 			3lp-gdb19   & 11  & 55  & 60.0(0.00)$\dagger$\cellcolor{lightgray}     & 60.0    & 60.0(0.00)\cellcolor{lightgray}    & 60.0   & 64.6(5.12)$\ddagger$  & 55.0    \\
 			3lp-gdb20   & 22  & 121 & 129.4(3.92)$\ddagger$                        & 124.0   & 125.6(3.85)\cellcolor{lightgray}   & 121.0  & 131.6(2.11)$\ddagger$  & 127.0   \\
 			3lp-gdb21   & 33  & 156 & 258.2(11.32)$\dagger$\cellcolor{lightgray}   & 239.5  & 263.5(10.20)\cellcolor{lightgray}  & 246.0  & 257.9(8.13)$\dagger$\cellcolor{lightgray}  & 241.1 \\
 			3lp-gdb22   & 44  & 200 & 317.7 (9.57)                                 & 298.5  & 323.8(9.19)   & 302.9  & 307.9(1.13)\cellcolor{lightgray}  & 302.9 \\
 			3lp-gdb23   & 55  & 233 & 346.8(8.51)$\dagger$                         & 335.6  & 349.1(12.44)  & 328.4  & 337.4(0.00)\cellcolor{lightgray}     & 337.4 \\
 			w-d-l & -   & -   & 3-18-2                                       & -       &-                 & -      & 14-2-7         & -      \\
 			No.best     & -   & -   & -                                   & 11      & -               & 10     & -             & 7     \\
 			Ave.PDR     & -   & -   & 37.38\%                                      & -       & 37.26\%         & -      & 40.29\%         & -      \\
 			\bottomrule    
 		\end{tabular}
 		
 	}
 \end{table*}

\begin{table*}[!htb]
	\scriptsize
	\caption{Results on the $3lp$-$egl$ benchmark test set  in terms of costs of solutions given a predefined time budget. For each instance, the average performance of an algorithm is indicated with a dark background if it is the best among all comparison algorithms. ``$\ddagger$" and ``$\dagger$" indicate KGMA-GN performs significantly better than and equivalently to the corresponding algorithm based on 20 independent runs according to the Wilcoxon rank-sum test with significant level $ p=0.05 $, respectively. }
	\label{tab:8}
	\centering
	\noindent \makebox[\textwidth][c]
	{
		\begin{tabular}{@{}ccccccccccccc@{}}
			\toprule
			
			Instance  & $|R|$ & LB    & \multicolumn{2}{c}{MAENS-GN}           & \multicolumn{2}{c}{KGMA-GN} & \multicolumn{2}{c}{VND-GN}    \\ 
			\cmidrule(lr){4-5} \cmidrule(lr){6-7} \cmidrule(lr){8-9}
			&     &                    & Ave(Std)                & Best                & Ave(Std)       & Best    & Ave(Std)    & Best    \\
			\midrule
			3lp-egl-e1-A & 51  & 3548  & 3588.2(7.59)$\dagger$ \cellcolor{lightgray}  & 3569.3    & 3591.6(7.74)\cellcolor{lightgray}   & 3573.5  & 3855.4(32.97)$\ddagger$  & 3726.9  \\
			3lp-egl-e1-B & 51  & 4498  & 4582.2(42.20)$\dagger$ \cellcolor{lightgray}  & 4498.0       & 4564.8(47.07)\cellcolor{lightgray}  & 4498.0    & 4710.8(51.14)$\ddagger$  & 4598.0    \\
			3lp-egl-e1-C & 51  & 5595  & 5895.6(62.65)$\dagger$ \cellcolor{lightgray}  & 5737.8    & 5870.4(59.40)\cellcolor{lightgray}   & 5746.7  & 6099.3(18.85)$\ddagger$  & 6043.9  \\
			3lp-egl-e2-A & 72  & 5018  & 5056.4(50.04)$\ddagger$  & 5018.0       & 5021.7(6.01)\cellcolor{lightgray}   & 5018.0    & 5180.0(0.00)$\ddagger$      & 5180.0    \\
			3lp-egl-e2-B & 72  & 6317  & 6377.4(22.60)$\dagger$ \cellcolor{lightgray}  & 6350.0      & 6361.6(13.97)\cellcolor{lightgray}  & 6350.0    & 6576.6(43.85)$\ddagger$  & 6457.0    \\
			3lp-egl-e2-C & 72  & 8335  & 8500.3(98.31)$\dagger$ \cellcolor{lightgray} & 8365.5    & 8467.5(70.71)\cellcolor{lightgray}  & 8368.0    & 8692.8(27.03)$\ddagger$  & 8575.0    \\
			3lp-egl-e3-A & 87  & 5898  & 6091.9(99.65)$\ddagger$  & 5924.0      & 6032.8(133.94)\cellcolor{lightgray} & 5898.0    & 6357.6(100.19)$\ddagger$ & 6089.0    \\
			3lp-egl-e3-B & 87  & 7744  & 8285.4(137.10)$\dagger$ \cellcolor{lightgray} & 8077.6    & 8271.1(123.09)\cellcolor{lightgray} & 7947.2  & 8802.7(73.06)$\ddagger$  & 8510.5  \\
			3lp-egl-e3-C & 87  & 10244 & 10404.0(61.50)$\ddagger$   & 10345.0     & 10346.6(16.49)\cellcolor{lightgray}  & 10326.0   & 10731.8(59.48)$\ddagger$  & 10605.0   \\
			3lp-egl-e4-A & 98  & 6408  & 7036.9(112.66)$\dagger$ \cellcolor{lightgray} & 6839.3     & 7028.7(67.73)\cellcolor{lightgray}  & 6795.2  & 7212.9(5.26)$\ddagger$   & 7191.1  \\
			3lp-egl-e4-B & 98  & 8935  & 9190.4(57.97)$\ddagger$  & 9109.1    & 9140.6(44.32)\cellcolor{lightgray}  & 9083.9  & 9549.6(73.46)$\ddagger$  & 9393.0    \\
			3lp-egl-e4-C & 98  & 11512 & 12362.4(236.91)$\ddagger$ & 12088.7   & 12185.1(190.03)\cellcolor{lightgray} & 11936.1 & 12631.5(169.83)$\ddagger$ & 12161.7 \\
			3lp-egl-s1-A & 75  & 5018  & 5120.0(51.02)$\ddagger$  & 5058.0       & 5039.2(58.10)\cellcolor{lightgray}   & 5018.0    & 5388.7(57.15)$\ddagger$  & 5282.0    \\
			3lp-egl-s1-B & 75  & 6388  & 7080.4(247.18)$\ddagger$ & 6845.0      & 6968.4(200.33)\cellcolor{lightgray} & 6862.3  & 7502.9(235.68)$\ddagger$ & 6794.2  \\
			3lp-egl-s1-C & 75  & 8518  & 10635.6(292.52)$\dagger$\cellcolor{lightgray} & 10015.3   & 10573.9(215.55)\cellcolor{lightgray} & 10268.8 & 10729.3(375.96)$\dagger$\cellcolor{lightgray} & 9894.7  \\
			3lp-egl-s2-A & 147 & 9825  & 10643.2(97.65)$\dagger$ \cellcolor{lightgray} & 10469.6   & 10572.8(123.73)\cellcolor{lightgray} & 10341.6 & 11313.6(96.08)$\ddagger$  & 11171.4 \\
			3lp-egl-s2-B & 147 & 13017 & 13562.7(65.58)$\ddagger$  & 13464.0    & 13470.0(44.75)\cellcolor{lightgray}  & 13410.0   & 14071.0(0.00)$\ddagger$      & 14071.0   \\
			3lp-egl-s2-C & 147 & 16425 & 18521.3(402.58)$\dagger$ \cellcolor{lightgray} & 17700.4   & 18278.0(489.49)\cellcolor{lightgray} & 17488.8 & 18858.2(284.33)$\ddagger$ & 18401.3 \\
			3lp-egl-s3-A & 159 & 10165 & 10901.2(70.86)$\ddagger$  & 10756.0   & 10773.4(39.00)\cellcolor{lightgray}  & 10715.0   & 11524.5(71.13)$\ddagger$  & 11330.4 \\
			3lp-egl-s3-B & 159 & 13648 & 14139.8(53.12)$\ddagger$  & 14026.5   & 14096.4(53.34)\cellcolor{lightgray}  & 13978.5 & 14586.1(0.00)$\ddagger$      & 14586.1 \\
			3lp-egl-s3-C & 159 & 17188 & 18112.5(137.87)$\ddagger$ & 17847.0     & 17934.3(91.52)\cellcolor{lightgray}  & 17788.2 & 18953.8(132.69)$\ddagger$ & 18684.4 \\
			3lp-egl-s4-A & 190 & 12153 & 12831.1(106.00)$\ddagger$    & 12641.0    & 12713.2(83.84)\cellcolor{lightgray}  & 12612.0   & 13355.0(0.00)$\ddagger$      & 13355.0   \\
			3lp-egl-s4-B & 190 & 16113 & 16858.3(123.26)$\ddagger$ & 16673.0    & 16655.6(68.91)\cellcolor{lightgray}  & 16524.0   & 17609.6(62.21)$\ddagger$  & 17446.0   \\
			3lp-egl-s4-C & 190 & 20430 & 21281.8(88.73)$\ddagger$  & 21083.0    & 21039.1(83.19)  & 20918.8 & 20920.4(0.00)\cellcolor{lightgray}      & 20920.4 \\
			w-d-l  & -   & -     & 14-10-0                   & -          &-                & -       & 22-1-1           & -        \\
			No.best      & -   & -     & -                         & 6          & -             & 19      & -              & 2       \\
			Ave.PDR      & -   & -     & 5.73\%                    & -          & 4.94\%         & -       & 9.60\%          & -        \\ 
			\bottomrule
		\end{tabular}
		
	}
\end{table*}


\begin{table*}[]
	\scriptsize
	\caption{Results on the $Solomon$-25 benchmark test set  in terms of costs of solutions given a predefined time budget (except for CG). For each instance, the average performance of an algorithm is indicated with a dark background if it is the best among all comparison algorithms. ``$\ddagger$" and ``$\dagger$" indicate KGMA-GN performs significantly better than and equivalently to the corresponding algorithm based on 20 independent runs according to the Wilcoxon rank-sum test with significant level $ p=0.05 $, respectively. }
	\label{tab:10}
	\centering
	\noindent \makebox[\textwidth][c]
	{
		\begin{tabular}{@{}ccccccccccc@{}}
			\toprule
			Instance    & \multicolumn{2}{c}{CG}         & \multicolumn{3}{c}{MAENS-GN}      & \multicolumn{2}{c}{KGMA-GN} & \multicolumn{2}{c}{VND-GN} \\
				\cmidrule(lr){2-3} \cmidrule(lr){4-6} \cmidrule(lr){7-8} \cmidrule(lr){9-10}
			  	& Cost  & Time  & Ave(Std)     & Best   & Time   & Ave(Std)           & Best   & Ave(Std)    & Best    \\
			\midrule
			r101    & 893.7 & 258.4    & 1014.3(10.84)$\dagger$\cellcolor{lightgray}& 1003.4 & 16.900 & 1015.3(11.93)\cellcolor{lightgray}  & 1001.0 & 1051.0(1.82)$\ddagger$   & 1050.3  \\
			r102    & 806.0 & 410.0     & 865.6(11.78)$\ddagger$ & 853.8  & 16.992 & 864.0(6.73)\cellcolor{lightgray}    & 853.3  & 900.8(0.00)$\ddagger$      & 900.8   \\
			r103     & 726.1 & 80.5    & 770.7(2.15)     \cellcolor{lightgray}       & 767.0  & 16.990 & 771.8(1.20)    & 768.8  & 807.5(3.31)$\ddagger$   & 806.8   \\
			r104   & 690.0 & 571.1     & 715.2(8.00)$\dagger$ \cellcolor{lightgray} & 703.0  & 12.491 & 711.3(5.46) \cellcolor{lightgray}   & 705.2  & 736.8(0.00)$\ddagger$      & 736.8   \\
			r105    & 780.4 & 118.2    & 839.7(3.16)$\dagger$ \cellcolor{lightgray} & 836.3  & 14.870 & 856.5(20.17)\cellcolor{lightgray}   & 836.3  & 870.8(24.84)$\ddagger$  & 853.9   \\
			r106    & 721.4 & 236.9     & 744.0(4.62)$\dagger$\cellcolor{lightgray}  & 740.1  & 17.315 & 743.7(5.84) \cellcolor{lightgray}   & 740.1  & 838.6(14.22)$\ddagger$  & 820.7   \\
			r107   & 679.5 & 137.8    & 700.5(2.81)$\dagger$ \cellcolor{lightgray} & 697.5  & 17.617 & 699.9(2.77)\cellcolor{lightgray}    & 697.5  & 742.4(7.30)$\ddagger$    & 740.1   \\
			r108    & 647.7 & 232.9     & 663.1(10.19)$\dagger$\cellcolor{lightgray} & 653.1  & 13.093 & 662.5(8.06)\cellcolor{lightgray}    & 655.6  & 691.5(4.38)$\ddagger$   & 689.4   \\
			r109    & 691.3 & 71.7     & 744.0(2.33)$\dagger$ \cellcolor{lightgray} & 734.7  & 16.114 & 744.5(4.13)\cellcolor{lightgray}    & 734.7  & 777.8(0.00)$\ddagger$      & 777.8   \\
			r110   & 668.8 & 125.9     & 716.9(10.05)$\dagger$\cellcolor{lightgray} & 697.2  & 17.043 & 718.8(10.00)\cellcolor{lightgray}   & 697.2  & 730.0(0.68)$\ddagger$   & 729.8   \\
			r111   & 676.3 & 157.8     & 710.5(5.16)$\ddagger$  & 707.9  & 18.448 & 707.9(0.00)\cellcolor{lightgray}    & 707.9  & 752.8(10.38)$\ddagger$  & 746.3   \\
			r112  & 643.0 & 356.1      & 666.3(5.17)$\dagger$ \cellcolor{lightgray} & 656.0  & 17.774 & 665.5(6.93) \cellcolor{lightgray}   & 660.3  & 698.1(5.72)$\ddagger$   & 696.1   \\
			rc101   & 623.8 & 62.4     & 648.4(3.06)$\dagger$ \cellcolor{lightgray} & 643.9  & 20.898 & 649.0(1.07)\cellcolor{lightgray}    & 645.0  & 681.7(0.00)$\ddagger$      & 681.7   \\
			rc102   & 598.3 & 179.8    & 611.4(5.07)$\dagger$ \cellcolor{lightgray} & 606.6  & 21.458 & 611.0(3.57)\cellcolor{lightgray}    & 606.6  & 770.9(13.81)$\ddagger$  & 761.5   \\
			rc103   & 585.1 & 941.5    & 607.8(4.95)$\dagger$ \cellcolor{lightgray} & 598.1  & 22.353 & 607.1(3.74)\cellcolor{lightgray}    & 600.7  & 645.8(0.85)$\ddagger$   & 645.4   \\
			rc104   & 572.4 & 3183.1   & 603.3(3.53)$\ddagger$  & 595.4  & 23.169 & 599.8(2.83)\cellcolor{lightgray}    & 595.0  & 609.0(0.00)$\ddagger$      & 609.0     \\
			rc105  & 623.1 & 743.7     & 656.3(9.23)$\ddagger$  & 641.1  & 23.280 & 647.8(3.19)\cellcolor{lightgray}    & 644.6  & 679.1(3.51)$\ddagger$   & 677.3   \\
			rc106   & 588.7 & 372.6    & 605.6(0.00)    \cellcolor{lightgray}        & 605.6  & 20.319 & 606.8(2.05)    & 605.6  & 631.3(0.00)$\ddagger$      & 631.3   \\
			rc107  & 548.3 & 824.8    & 567.0(2.15)$\dagger$\cellcolor{lightgray}  & 565.4  & 13.716 & 567.4(2.32)\cellcolor{lightgray}    & 565.4  & 608.9(9.54)$\ddagger$   & 600.1   \\
			rc108   & 544.5 & 3457.1   & 553.5(4.41)$\dagger$\cellcolor{lightgray}  & 546.8  & 13.846 & 555.5(3.02) \cellcolor{lightgray}   & 550.7  & 564.8(8.93)$\ddagger$   & 555.9   \\
			w-d-l     &0-0-20 &-      & 4-14-2       & -      & -      &  -               & -      & 20-0-0           &  -       \\
			No.best    &  - &  -     &  -            & 17     & -      & -               & 12     & -              & 9       \\
			Ave.PDR    &  - &  -      & 4.95\%       & -      & -      & 4.95\%          & -      & 10.93\%           &  -    \\
			Ave.Time    & -   & 626.115 & -         & -     & 17.734           & -            & -     & - & -        \\
			\bottomrule  
		\end{tabular}
		
	}
\end{table*}

\subsubsection{Benchmark Test Sets}
\label{testset}
 Two types of benchmark test sets are commonly used. The first \del{is the} \rev{category consists of test} sets with two-segment linear functions, namely $ gdb $ in  2LP (called $2lp$-$ gdb $ here) \cite{tagmouti2010variable} and  $ egl $ in  2LP (called 2$lp$-$ egl $ here) \cite{tagmouti2010variable}. The second \del{is the} \rev{category consists of test}  sets with three-segment linear functions, namely $ gdb $ in  3LP (called $3lp$-$gdb $ here) \cite{li2026novel}, $ egl $ in  3LP (called 3$lp$-$ egl $ here) \cite{li2026novel} and  $ Solomon $-25 \cite{tagmouti2007arc}. 
Among them,
  $2lp$-$gdb$ and $3lp$-$gdb$  are adapted from the static CARP test set $gdb$ \cite{golden1983computational}, and therefore they are small-scale test sets with up to 55 tasks. Similarly, $2lp$-$egl$ and $3lp$-$egl$ are adapted from the static CARP benchmark set $egl$ \cite{eglese1994routeing}. $egl$  is designed based on the winter gritting application in Lancashire, UK, and it is a larger-scale benchmark set with up to 190 tasks. Hence, $2lp$-$egl$ and $3lp$-$egl$ are also larger-scale test sets with a maximum of 190 tasks. Finally, $Solomon$-25  is derived from $Solomon$ \cite{solomon1987algorithms}, which is a VRP benchmark set with time windows.  $Solomon$-25 is a small-scale test set containing 25 tasks.


\subsubsection{Compared Algorithms}
\label{comalgo}
 For  test sets in 2LP, VND \cite{tagmouti2010variable} and MAENS-GN \cite{li2026novel}  are  selected as baseline algorithms for comparison.   
 Since vehicle departure times are not considered in 2LP,  only the first stage of MAENS-GN (i.e., MAENS) and  KGMA-GN (i.e., KGMA) is executed. 
 
  For  test sets in 3LP,  CG \cite{tagmouti2007arc}, VND-GN \cite{li2026novel} and MAENS-GN are compared with KGMA-GN.
  As an exact algorithm, CG has results only on $Solomon$-25 and is evaluated only on that set. 
VND-GN \cite{li2026novel} is the variant of VND \cite{tagmouti2010variable}, adapted to solve the 3LP. 
\subsubsection{Performance Metrics}
\label{metric}
The performance metrics include  solution quality and optimization speed. The solution quality is measured by the cost of the final solution obtained given a predefined time budget, as defined in  (\ref{equ:1}). The optimization  speed  is measured by  the actual runtime (in seconds) required to achieve a predefined solution quality.
\subsubsection{Experimental Protocol}
\label{keypara}

The parameter settings for MAENS-GN follow those in \cite{li2026novel}.  VND (or VND-GN) adopts the initialization strategy from  \cite{li2026novel}, which  is   a stochastic algorithm.  The results of CG are taken directly from \cite{li2026novel}. 

 Key parameters of  KGMA-GN are listed as follows. The population size ($ psize $) and    local search  probability ($ Pls $) are set to 10 and 0.1, respectively,  consistent with those of  MAENS-GN.  Besides, for the experiments given a predefined  target solution quality, the number of generations ($gnum$) for  KGMA-GN is allowed for up to 600  if the target cannot be achieved. In  Criterion  1, the control parameter $ \lambda $ is set to 1, which provides a natural and interpretable decision boundary: Criterion 1 is triggered when the total time gap after a move exceeds that before the move.  Other parameters follow the settings in \cite{li2026novel}. 
 
 Each stochastic algorithm is independently run 20 times  on each instance. 
For  fairness, all stochastic algorithms are implemented in C++ and run on the same platform: Intel(R) Xeon(R) Platinum 9242 CPU @ 2.30GHz.

\subsection{Comparison  of Solution Quality Given a Predefined  Runtime}
\label{expe:qua}
The solution quality of these algorithms is measured given a  predefined time budget. Following \cite{li2026novel}, the  budget 
 is set to the time  required for MAENS-GN to complete 50 generations, as MAENS-GN generally has longer runtimes. 

Tables \ref{tab:2}-\ref{tab:10} show   the final solution costs of the compared algorithms on five test sets given a predefined  runtime (except for CG).  Due to space limitations, Tables \ref{tab:2} and \ref{tab:4} report only the overall comparison results across the benchmark test sets. The complete per-instance results are provided in the Supplementary Material (Tables S1 and S2).
 The  contents of each table are  briefly explained below.
 \begin{itemize}
	\item   “$|R|$” and “LB” represent the number of  tasks  and the lower bound \cite{li2026novel}, respectively.
	\item  “Ave(Std)” shows the average cost and standard deviation over 20 independent runs, and  “Best” shows the best cost achieved among those 20 runs.
	\item   In Table \ref{tab:10}, ``Time"  refers to the runtime for CG, while for other algorithms, ``Time"  refers to the average runtime over 20 independent runs.
	\item  “w-d-l”  indicates the number of instances  where the proposed algorithm is better than, not significantly different from, or worse than the corresponding  algorithm, based on average costs and  the Wilcoxon rank-sum test \cite{wilcoxon1992individual}. 
	\item  “No.best” refers to the number of instances   where the algorithm achieves the best cost among all compared algorithms. 
	\item  “Ave.PDR” shows the average performance degradation rate (PDR).  PDR = $(C_1-C_2)/C_2 \times  100\% $, where $ C_1 $   denotes   the solution cost  of the algorithm  and $ C_2 $  denotes the lower bound or the best cost among  all the algorithms. 
	\item  “Ave.Time”  represents the average runtime.
\end{itemize}



To ensure a  fair comparison with stochastic algorithms,  both the average and  best performance are considered. when compared with the exact algorithm (CG), both the final solution cost and runtime are  examined.

The first comparison with stochastic algorithms focuses on the average performance. In terms of Ave.PDR, KGMA-GN performs no worse than the other algorithms on all five test sets. Specifically, there are two interesting observations. The first is that the performance gap between KGMA-GN and the other algorithms is larger on larger-scale instances than on small-scale ones. This is because that small-scale instances are relatively simple, leaving less room for performance gaps. In contrast,  larger-scale instances are more complex, allowing KGMA-GN to better demonstrate its advantages.
The second observation is that the performance gap between KGMA-GN and VND (or VND-GN) is generally  greater than that between KGMA-GN and MAENS-GN. This is because MAENS-GN produces better solution quality than VND (or VND-GN). 
Further statistical tests (i.e., w-d-l) \del{shows} \rev{show}  that KGMA-GN still performs better than the other two algorithms on all five test sets, confirming  its superiority in average performance. 

The second comparison with stochastic algorithms focuses on the best performance. In terms of No.best, KGMA-GN is obviously better than VND (or VND-GN) on all five test sets. Compared with the same type of algorithm MAENS-GN, KGMA-GN performs better on the  test sets of 2LP (i.e., $2lp$-$gdb$ and $2lp$-$egl$) and the larger-scale test set of 3LP (i.e., $3lp$-$egl$), while it is slightly worse on the two small-scale test sets of 3LP (i.e., $3lp$-$gdb$ and $Solomon$-25).

As shown in Table \ref{tab:10},  KGMA-GN performs worse than the exact algorithm CG in terms of costs of solutions on $Solomon$-25, but KGMA-GN is much faster. Moreover, since CG requires significant time and resources to find optimal solutions, it can only handle small-scale instances \cite{tagmouti2007arc}. In contrast, KGMA-GN can efficiently handle larger-scale instances within  a limited time.

In summary, KGMA-GN achieves high solution quality given the predefined runtime. The better solution quality of KGMA-GN comes not only from KGIS, which  helps generate a good initial population, but more importantly from the significant time savings brought by filtering  out failed moves in KGSLSS. 

\begin{table}[]
	\scriptsize
	\caption{Runtime comparison results of KGMA-GN and MAENS-GN on all five test sets given a target solution quality. w-d-l indicates the number of instances on the test set where KGMA-GN has a shorter, not significantly different, or longer runtime  compared to MAENS-GN  given a target solution quality. The comparison is based on the average runtime from 20 independent runs and the Wilcoxon rank-sum test with significance level $p$=0.05.}
	\label{tab:arrivetime}
	\centering
	\begin{tabular}{@{}ccccccccc@{}}
		\toprule
		
      & $2lp$-$gdb$ & $2lp$-$egl$ & $3lp$-$gdb$ & $3lp$-$egl$ & $Solomon$-25 \\
	  \midrule
w-d-l & 13-8-2  & 19-5-0  & 10-7-6  & 22-1-1  & 11-6-3    
  
  \\ \bottomrule
	\end{tabular}
\end{table}

\begin{table}[!htb]
	\scriptsize
	\caption {Comparison results of KGMA-GN with traditional initialization  and KGMA-GN with KGIS on  all  five test sets in terms of costs of solutions. ``Init-best" and ``Best" denote the best value of the initialization strategy and the whole algorithm on each test set on 20 independent runs, respectively. For each test set, the value is indicated with a dark background if it is the best among all comparison algorithms.}
	\label{tab:11}
	\centering
	\noindent \makebox[\textwidth][l]
	{
		\begin{tabular}{ccccc}
			\toprule
			Test set  & \multicolumn{2}{c}{\makecell[c]{KGMA-GN \\ with  traditional initialization}} & \multicolumn{2}{c}{\makecell[c]{KGMA-GN \\ with KGIS}}         \\
			\cmidrule(lr){2-3}	\cmidrule(lr){4-5} 
			& Init-best          & Best        & Init-best       & Best             \\
			\midrule
			$2lp$-$ gdb  $   & 367.0                 & 256.3       & 323.7 \cellcolor{lightgray}  & 256.2 \cellcolor{lightgray}   \\
			$2lp$-$egl $    & 16763.2             & 10073.7     & 14915.8 \cellcolor{lightgray} & 10045.5 \cellcolor{lightgray}\\
			$3lp$-$gdb  $   & 21210.7             & 338.7       & 16671.3\cellcolor{lightgray} & 338.1   \cellcolor{lightgray}\\
			$3lp$-$egl $    & 261313.5            & 10133.5     & 15028.9\cellcolor{lightgray} & 10128.0 \cellcolor{lightgray}\\
			$ Solomon $-25 & 1514.1 \cellcolor{lightgray}    & 694.1       & 1666.0           & 692.4 \cellcolor{lightgray}\\  
			\bottomrule 
		\end{tabular}
		
	}
\end{table}
\subsection{Comparison  of Runtime Given a Target Solution Quality}
\label{expe:time}
Section \ref{expe:qua} shows that MAENS-GN and KGMA-GN perform better than the other  algorithms in terms of  solution quality on five test sets. To further evaluate their efficiency, the runtime required   to reach a target solution quality is compared.  The target solution quality is set as the average result  obtained by MAENS-GN over 20 independent runs on each instance over  50 generations \cite{li2026novel}. 

Table \ref{tab:arrivetime} shows the runtime comparison between KGMA-GN and MAENS-GN with statistical testing. As shown in  this table,  KGMA-GN achieves  shorter runtimes than MAENS-GN on all five test sets, especially on the larger-scale test sets $2lp$-$egl$ and $3lp$-$egl$. 

Overall, the results show that KGMA-GN generally  reaches the target solution quality much faster than MAENS-GN on all five benchmark test sets. These findings demonstrate that incorporating problem-specific knowledge effectively accelerates the search process, enabling KGMA-GN to achieve  the higher search efficiency.
\subsection{Effectiveness Test of KGIS}
\label{sec:inits}

In KGIS, the initial solution is constructed   by  minimizing the sum of the travel distance and problem-specific knowledge (Algorithm \ref{alg:1}). To verify the effectiveness of KGIS, we compare the KGMA-GN with KGIS to the KGMA-GN with  traditional  initialization (i.e., KGIS without knowledge). This section adopts the best solution cost as the performance metric, which is widely used in routing problems \cite{tang2009memetic, jin2020planning, liu2021memetic}. 
Both methods were run  for 50 generations on  five test sets,  with  20 independent runs for each.

\begin{table*}[!htb]
	\scriptsize
	\caption {Time comparison of traditional local search operators and knowledge-guided local search operators on all five test sets. The values in parentheses show how many times the traditional operator is slower than the  corresponding knowledge-guided one. For each test set, the average runtime of an algorithm is indicated with a dark background if it is the best among all comparison algorithms. The results are based on the average value on 20 independent runs.}
	\label{tab:12}
	\centering
	\noindent \makebox[\textwidth][c]
	{
		\begin{tabular}{>{\centering\arraybackslash}p{1.7cm}>{\centering\arraybackslash}p{2.2cm}>{\centering\arraybackslash}p{2.2cm}>{\centering\arraybackslash}p{1.8cm}>{\centering\arraybackslash}p{2.2cm}>{\centering\arraybackslash}p{2.2cm}>{\centering\arraybackslash}p{1.8cm}}
			\toprule
			Test set  & \multicolumn{3}{c}{Traditional local search operators} & \multicolumn{3}{c}{ Knowledge-guided local search operators } \\
			\cmidrule(lr){2-4}	\cmidrule(lr){5-7} 
			& single insertion     & double insertion     & swap        & single insertion    & double insertion    & swap              \\
			\midrule
			$2lp$-$ gdb $    & 1.768                & 2.466                & 1.190       & \cellcolor{lightgray} 0.785 (2.3)      & \cellcolor{lightgray} 0.685 (3.6)      & \cellcolor{lightgray}  0.103 (11.6)   \\
			$2lp$-$egl $    & 148.096              & 246.764              & 175.779     & \cellcolor{lightgray}  48.594 (3.0)    & \cellcolor{lightgray}  47.244 (5.2)     & \cellcolor{lightgray} 7.567 (23.2)    \\
			$3lp$-$ gdb $    & 4.270                & 5.566                & 3.622       & \cellcolor{lightgray} 1.632 (2.6)      & \cellcolor{lightgray} 1.334 (4.2)     &\cellcolor{lightgray} 0.353 (10.3)    \\
			$3lp$-$ egl $    & 128.845              & 213.507              & 149.745     &\cellcolor{lightgray} 46.292 (2.8)      & \cellcolor{lightgray} 46.311 (4.6)     & \cellcolor{lightgray} 8.970 (16.7)     \\
			$ Solomon $-25 & 4.850                & 8.018                & 5.865       &\cellcolor{lightgray} 1.772 (2.7)      &\cellcolor{lightgray} 1.636 (4.9)     & \cellcolor{lightgray} 0.432 (13.6)   \\
			\bottomrule 
		\end{tabular}
		 
	}
\end{table*}

\begin{figure*}[!htb]
		\centering
	
	\noindent \makebox[\textwidth][l]{
		\subfloat[\textbf{ $2lp$-$ gdb $}]{
			\begin{minipage}[t]{0.196\linewidth}
							\centering
				\hspace*{-0.5cm}
				\includegraphics[width=1.6in]{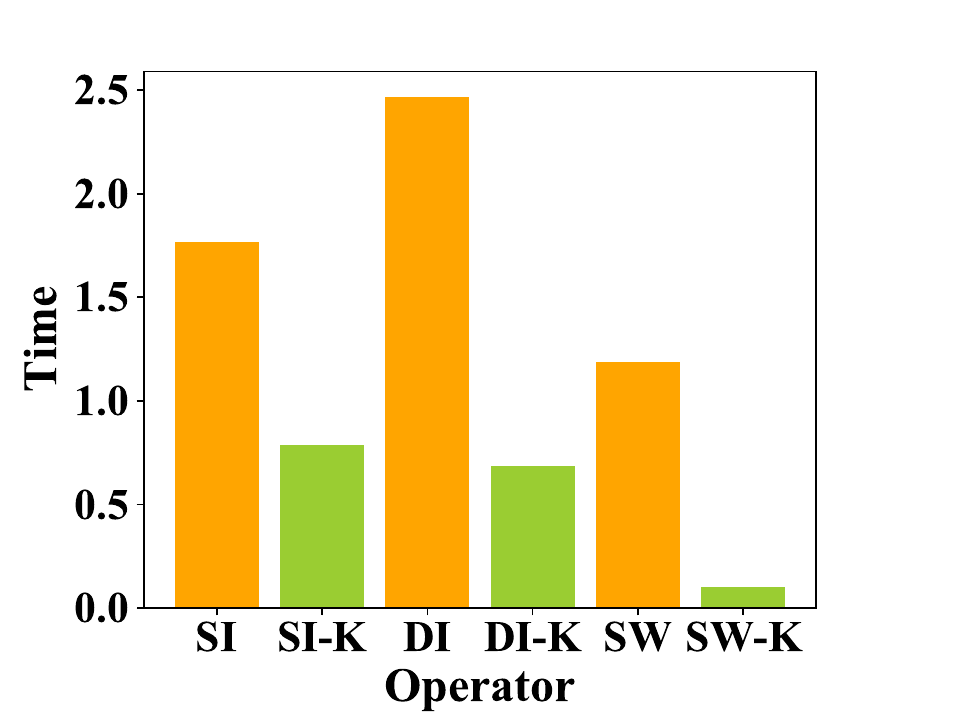}
				
			\end{minipage}%
		}%
		
		\subfloat[ \textbf{$2lp$-$ egl $}]{
			\begin{minipage}[t]{0.197\linewidth}
				\centering
				\hspace*{-0.5cm}
				\includegraphics[width=1.6in]{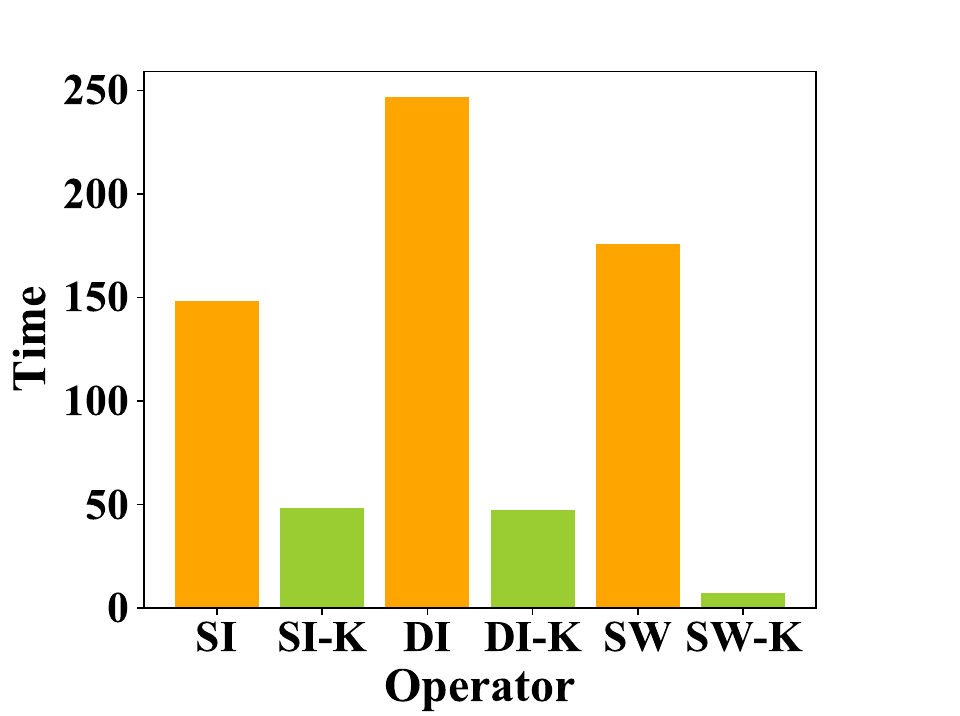}
				
			\end{minipage}%
		}%
		
		\subfloat[ \textbf{$3lp$-$ gdb $}]{
			\begin{minipage}[t]{0.196\linewidth}
				\centering
				\hspace*{-0.5cm}
				\includegraphics[width=1.6in]{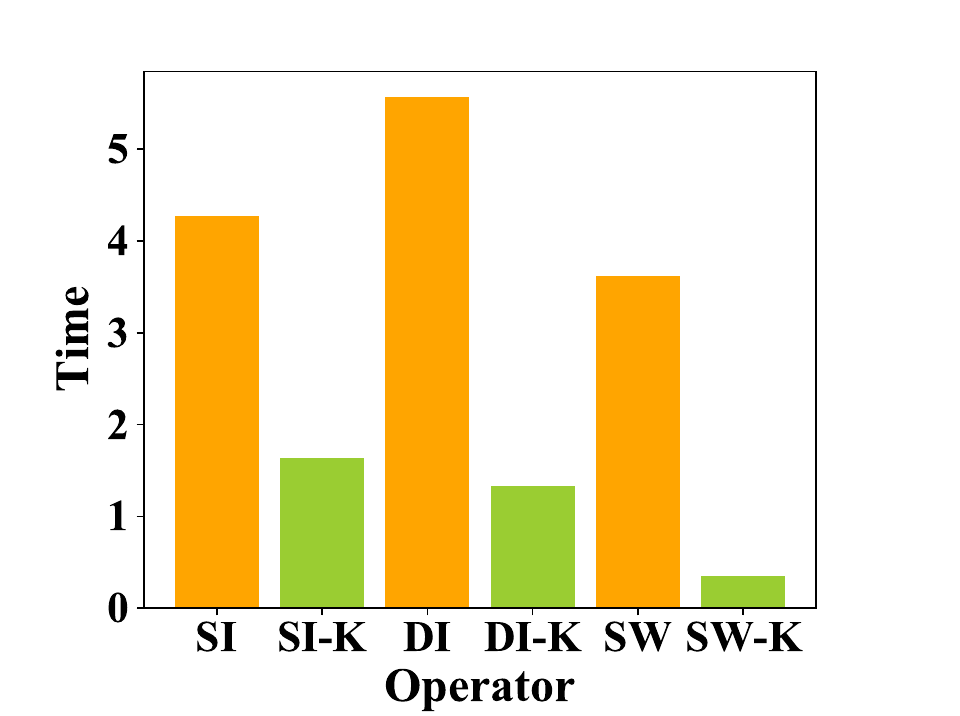}
				
			\end{minipage}%
		}%
		\subfloat[ \textbf{$3lp$-$ egl $}]{
			\begin{minipage}[t]{0.196\linewidth}
				\centering
				\hspace*{-0.5cm}
				\includegraphics[width=1.6in]{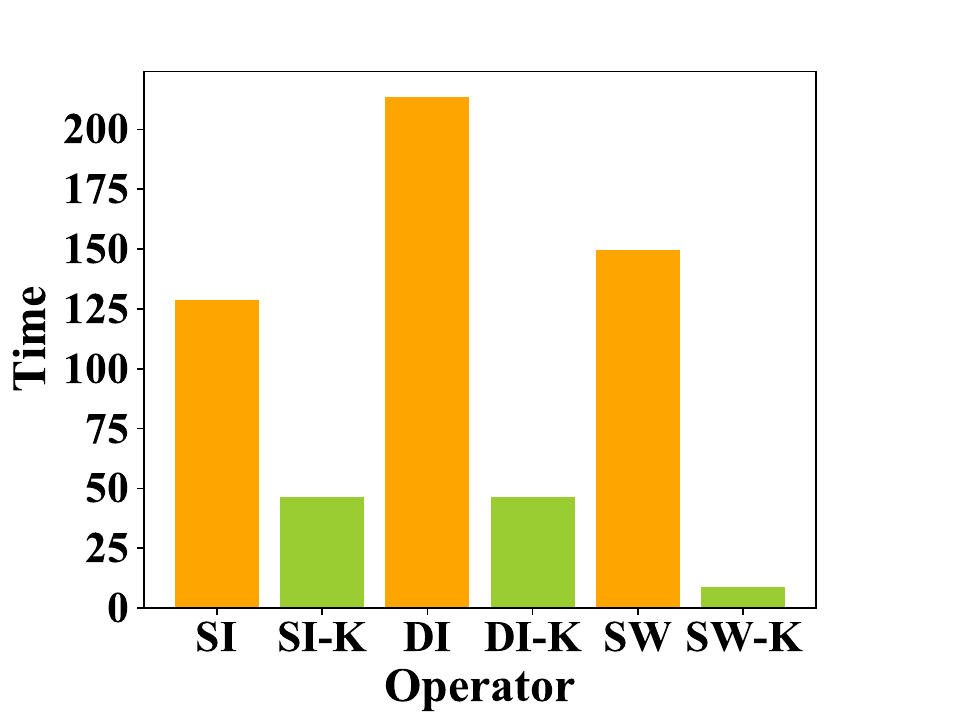}
				
			\end{minipage}%
		}%
		
		\subfloat[ \textbf{$ Solomon $-25}]{
			\begin{minipage}[t]{0.196\linewidth}
				\centering
				\hspace*{-0.5cm}
				\includegraphics[width=1.6in]{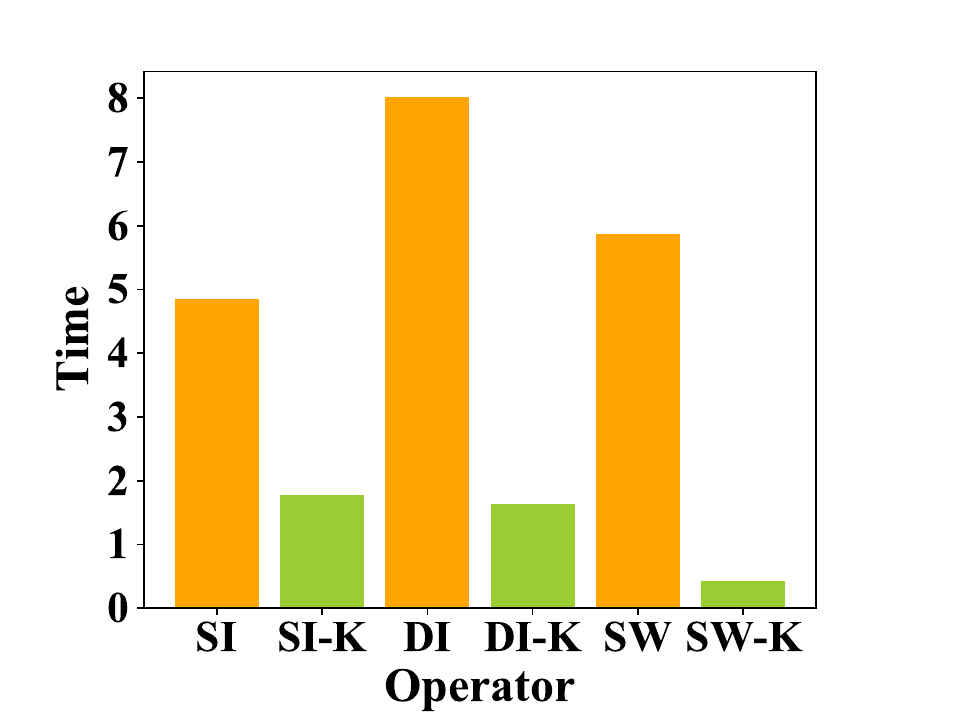}
				
			\end{minipage}%
		}%
	}
	
	\centering
	\caption{The time comparison of the traditional local search operators and the knowledge-guided local search operators on five test sets. ``SI", ``SI-K", ``DI", ``DI-K", ``SW" and ``SW-K" represent single insertion, knowledge-guided single insertion, double insertion, knowledge-guided double insertion, swap, and knowledge-guided swap, respectively.}
	\label{fig:loal}
\end{figure*}

The  results are presented  in Table \ref{tab:11}, where  KGMA-GN with KGIS is significantly better than KGMA-GN with traditional initialization on four test sets in terms of ``Init-best", and achieves better performance on all five test sets in terms of ``Best". 
Specifically, there are two observations.
First, the performance gap between the two algorithms in ``Init-best" is notably  larger than that in ``Best", showing that KGIS produces much better initial solutions than the traditional method. However,  this gap narrows as the number of iterations increases. 
Second, the performance gap  between the two algorithms is larger on larger-scale test sets than on small-scale ones.  This is because  small-scale instances are relatively simple, and both algorithms can achieve high-quality solutions, or even  optimal solutions, resulting in a small performance gap. In contrast,  larger-scale instances are  more challenging, which leads to a bigger performance gap. These results  confirm that KGIS is effective in most cases.

Overall, the results indicate that KGIS effectively improves the quality of the initial population, particularly on larger-scale instances. These improvements provide a higher starting point for the subsequent memetic search, thereby contributing to the superior performance of KGMA-GN.

\subsection{Effectiveness Test  of KGSLSS}
\label{sec:lso}
This section evaluates whether the knowledge-guided small-step-size local search operators in KGSLSS can significantly reduce runtime compared to the traditional ones.

Table \ref{tab:12} and Figure \ref{fig:loal} present  the runtime comparison between  traditional and knowledge-guided small-step-size local search operators on five test sets. Each type of operator is separately embedded into KGMA-GN, and the results are obtained after running 50 generations and 20 independent runs. 
The results show  that the knowledge-guided  operators significantly reduce the search time compared to the traditional ones on all five test sets, especially for the knowledge-guided swap operator, where the time is reduced by more than 10 times on each test set.

Overall, the results demonstrate that incorporating problem-specific knowledge into local search can substantially reduce runtime,  thereby   improving computational efficiency.
\section{Conclusion and future work}
\label{sec:conc}

This paper presents an efficient algorithmic solution KGMA-GN to a  computationally challenging problem CARPTDSC. Two knowledge-guided strategies (i.e., KGIS and KGSLSS) are designed to improve efficiency. 
KGIS builds a high-quality initial population and \del{speed} \rev{speeds} up convergence. KGSLSS uses problem-specific knowledge to quickly filter out failed moves, reducing  unnecessary evaluations and saving significant computation time. Experiments on five test sets demonstrate that KGMA-GN significantly  improves search efficiency. Two ablation studies further validate  the effectiveness of two proposed knowledge-guided strategies. 

Although KGMA-GN demonstrates good search efficiency, there is still room for improvement. First, the time-dependent function used in this work is relatively simple, and more realistic functions will be explored in the future. In addition, the search efficiency can be further improved. For example, the current search operators are applied in a fixed order, which may limit efficiency. Learning-based methods \cite{guo2025learning, wang2024deep} or large language models \cite{liu2024large}  could be used to capture the characteristics of the operators and adaptively determine their order, thereby further enhancing the overall efficiency. 

\begin{IEEEbiography}[{\includegraphics[width=1in,height=1.25in,clip,keepaspectratio]{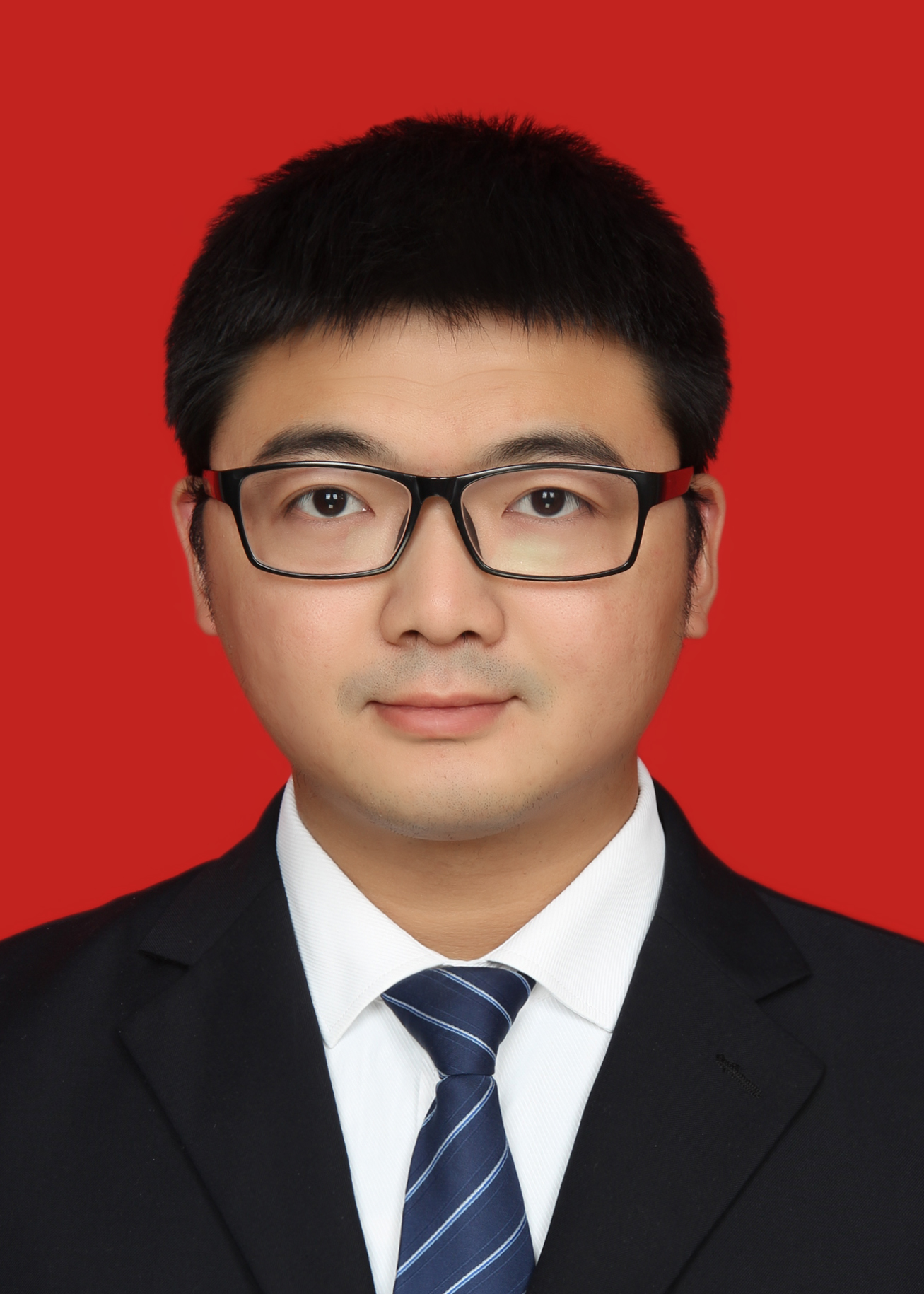}}]{Qingya Li}
received the B.S. and M.S. degrees from the Department of Computer Science, Xiangtan University, China, in 2015 and 2018, respectively. He is currently pursuing the Ph.D. degree with the Department of Computer Science and Engineering, Southern University of Science and Technology, China. His research interests include evolutionary computation and the capacitated arc routing problems.
\end{IEEEbiography}
\begin{IEEEbiography}[{\includegraphics[width=1in,height=1.25in,clip,keepaspectratio]{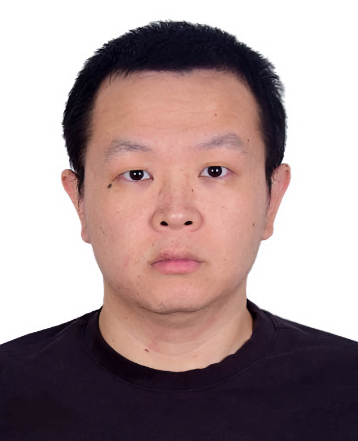}}]{Shengcai Liu} (Member, IEEE) received the B.Sc. degree in 2014 and the Ph.D. degree in 2020 from the University of Science and Technology of China, Hefei, China. He is currently an Assistant Professor with the Department of Computer Science and Engineering, Southern University of Science and Technology, Shenzhen, China. He is mainly interested in Learning to Optimize and has published more than 30 papers in top-tier refereed international conferences and journals. He is also an associate editor of the IEEE Transactions on Emerging Topics in Computational Intelligence.
\end{IEEEbiography}

\begin{IEEEbiography}[{\includegraphics[width=1in,height=1.25in,clip,keepaspectratio]{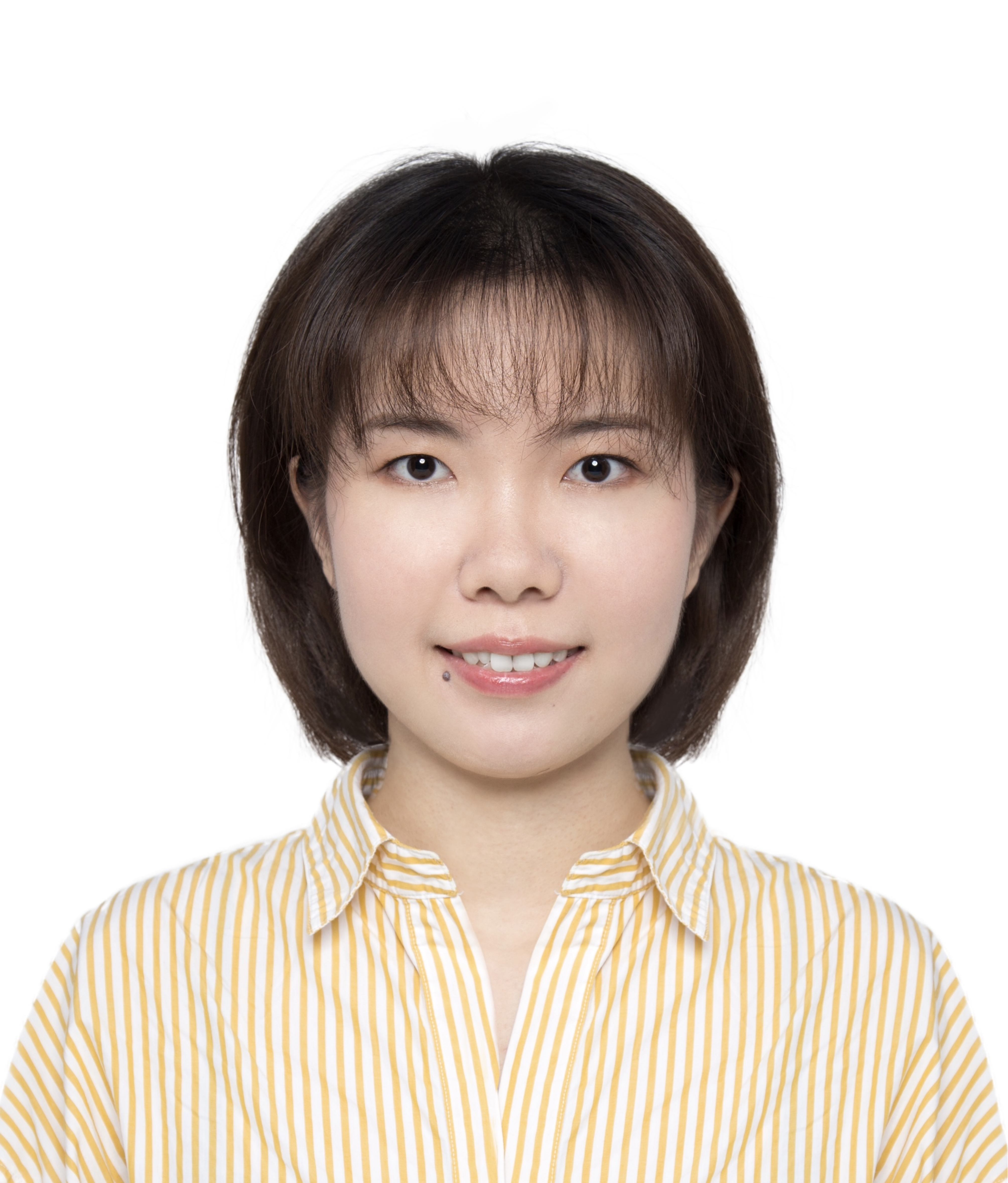}}]{Wenjie Chen} received the B.S. degree in automation from the Huazhong University of Science and Engineering, Wuhan, China, in 2016, and Ph.D. degree in systems engineering and engineering management from the City University of Hong Kong, Hong Kong, in 2020. She is currently an Assistant Professor with the School of Information Management, Central China Normal University, Wuhan, China. Her research is devoted to simulation optimization, machine learning, and real-world applications such as healthcare management and social network analysis.
\end{IEEEbiography}

\begin{IEEEbiography}[{\includegraphics[width=1in,height=1.25in,clip,keepaspectratio]{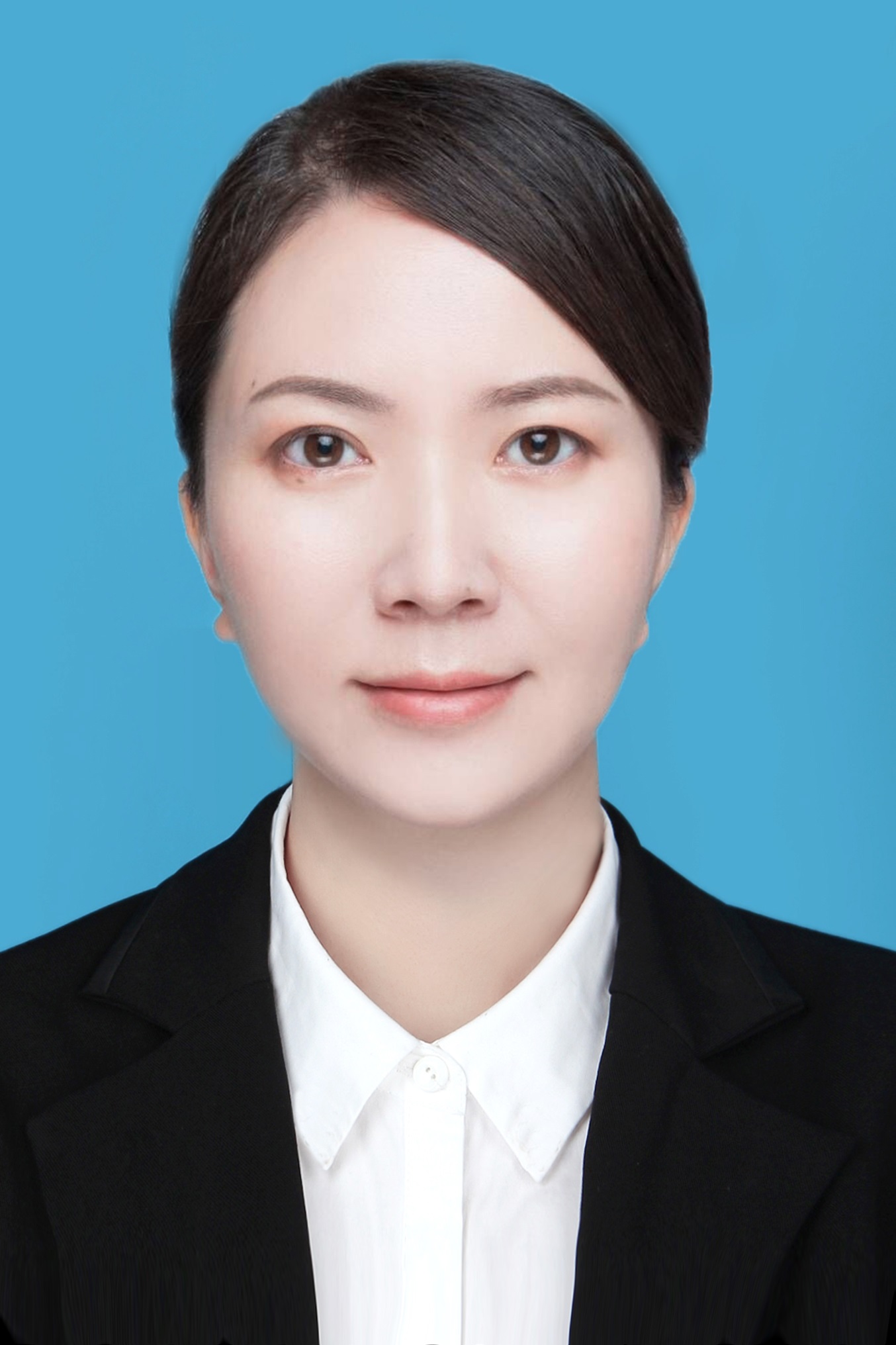}}]{Juan Zou} received the B.Sc. degree in computer applications technology and the M.Sc. degree in computer science and engineering from Xiangtan University of Science and Technology, Xiangtan, China, in 2001 and 2005, respectively, and the Ph.D. degree in mathematics from Xiangtan University, Xiangtan, in 2014.
She is currently a Professor of School of Artificial Intelligence and Statistics, Xiangtan University. She has over 50 publications and one monograph. Her current research interests include multiobjective optimization, swarm intelligence, artificial neural networks, and related applications.
\end{IEEEbiography}

\begin{IEEEbiography}[{\includegraphics[width=1in,height=1.25in,clip,keepaspectratio]{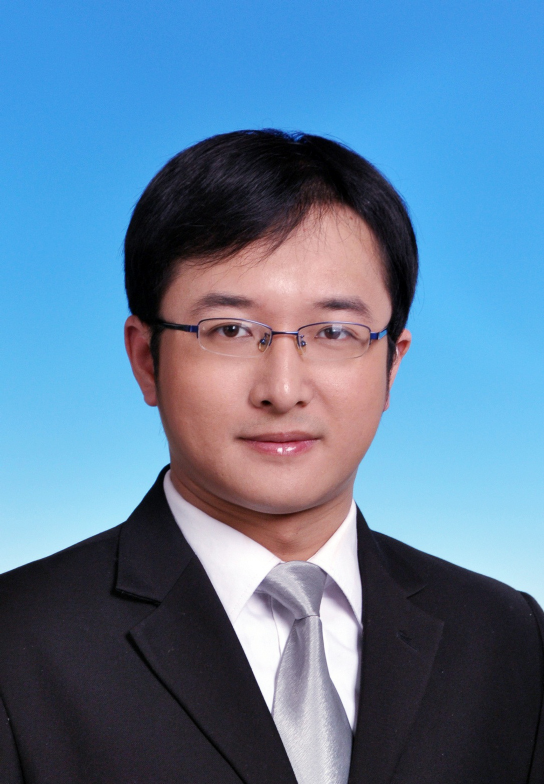}}]{Ke Tang} (Fellow, IEEE) received the Ph.D. degree in computer science from Nanyang Technological University, Singapore, in 2007. He is currently a chair professor at the Department of Computer Science and Engineering, at Southern University of Science and Technology (SUSTech), Shenzhen, China. His research interests mainly include evolutionary computation, machine learning, and their applications. He is a Fellow of IEEE and he was awarded the Newton Advanced Fellowship (Royal Society) and the Changjiang Professorship Ministry of Education (MOE) of China. He was the recipient of the IEEE Computational Intelligence Society Outstanding Early Career Award and the Natural Science Award of MOE of China. He has been actively involved in the research community by serving as an organizer/co-organizer of numerous conferences, an associate editor of the IEEE TRANSACTIONS ON EVOLUTIONARY COMPUTATION, and as a member of editorial boards for a few other journals.
\end{IEEEbiography}


\begin{IEEEbiography}[{\includegraphics[width=1in,height=1.25in,clip,keepaspectratio]{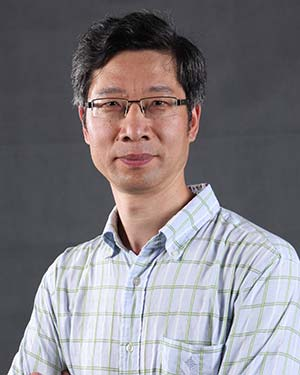}}]{Xin Yao} (Fellow, IEEE) received the B.Sc. degree from the University of Science and Technology of China (USTC) in 1982, the M.Sc. degree from the North China Institute of Computing Technologies in 1985, and the Ph.D. degree from USTC in 1990. He is the Tong Tin Sun Chair Professor of Machine Learning at Lingnan University, Hong Kong, China. He served as the president (2014-15) of IEEE CIS and the editor-in-chief (2003-08) of \textit{IEEE Transactions on Evolutionary Computation}. His major research interests include evolutionary computation, machine learning, and their real-world applications. His work won the 2001 IEEE Donald G. Fink Prize Paper Award; 2010, 2016 and 2017 IEEE Transactions on Evolutionary Computation Outstanding Paper Awards; 2011 IEEE Transactions on Neural Networks Outstanding Paper Award; 2010 BT Gordon Radley Award for Best Author of Innovation (Finalist); and other best paper awards at conferences. He received the 2012 Royal Society Wolfson Research Merit Award, the 2013 IEEE CIS Evolutionary Computation Pioneer Award and the 2020 IEEE Frank Rosenblatt Award.
\end{IEEEbiography}




\begin{thebibliography}{10}
\bibliographystyle{IEEEtran}
\bibitem{golden1981capacitated}
B.~L. Golden and R.~T. Wong, ``Capacitated arc routing problems,''
  \emph{Networks}, vol.~11, no.~3, pp. 305--315, 1981.

\bibitem{jia2025neural}
\rev{Y.-H. Jia, Q.~Zheng, Y.~Wang, Y.~Mei, W.-N. Chen, and Z.~Lin, ``A neural solver
  with traversal-based feature representation and adjacent attention for
  capacitated arc routing problem,'' \emph{IEEE Transactions on Intelligent
  Transportation Systems}, vol.~26, no.~12, pp. 22\,329--22\,343, 2025.}

\bibitem{zhang2023rocash2}
Y.~Zhang, Y.~Mei, H.~Zhang, Q.~Cai, and H.~Wu, ``RoCaSH2: An effective route
  clustering and search heuristic for large-scale multi-depot capacitated arc
  routing problem,'' \emph{IEEE Computational Intelligence Magazine}, vol.~18,
  no.~4, pp. 43--56, 2023.



\bibitem{tang2009memetic}
K.~Tang, Y.~Mei, and X.~Yao, ``Memetic algorithm with extended neighborhood
  search for capacitated arc routing problems,'' \emph{IEEE Transactions on
  Evolutionary Computation}, vol.~13, no.~5, pp. 1151--1166, 2009.

\bibitem{jin2020planning}
X.~Jin, H.~Qin, Z.~Zhang, M.~Zhou, and J.~Wang, ``Planning of garbage
  collection service: An arc-routing problem with time-dependent penalty
  cost,'' \emph{IEEE Transactions on Intelligent Transportation Systems},
  vol.~22, no.~5, pp. 2692--2705, 2020.

\bibitem{bodin1978computer}
L.~D. Bodin and S.~J. Kursh, ``A computer-assisted system for the routing and
  scheduling of street sweepers,'' \emph{Operations Research}, vol.~26, no.~4,
  pp. 525--537, 1978.

\bibitem{stern1979routing}
H.~I. Stern and M.~Dror, ``Routing electric meter readers,'' \emph{Computers \&
  Operations Research}, vol.~6, no.~4, pp. 209--223, 1979.

\bibitem{tagmouti2007arc}
M.~Tagmouti, M.~Gendreau, and J.-Y. Potvin, ``Arc routing problems with
  time-dependent service costs,'' \emph{European Journal of Operational
  Research}, vol. 181, no.~1, pp. 30--39, 2007.

\bibitem{tagmouti2010variable}
M.~Tagmouti, M.~Gendreau, and J.-Y. Potvin, ``A variable neighborhood descent heuristic for arc routing problems
  with time-dependent service costs,'' \emph{Computers \& Industrial
  Engineering}, vol.~59, no.~4, pp. 954--963, 2010.

\bibitem{tagmouti2011dynamic}
M.~Tagmouti, M.~Gendreau, and J.-Y. Potvin, ``A dynamic capacitated arc routing problem with time-dependent service
  costs,'' \emph{Transportation Research Part C: Emerging Technologies},
  vol.~19, no.~1, pp. 20--28, 2011.

\bibitem{arnold2019knowledge}
F.~Arnold and K.~S{\"o}rensen, ``Knowledge-guided local search for the vehicle
  routing problem,'' \emph{Computers \& Operations Research}, vol. 105, pp.
  32--46, 2019.

\bibitem{zhang2024knowledge}
H.~Zhang, Q.~Li, and X.~Yao, ``Knowledge-guided optimization for complex
  vehicle routing with 3{D} loading constraints,'' in \emph{International
  Conference on Parallel Problem Solving from Nature}.\hskip 1em plus 0.5em
  minus 0.4em\relax Springer, 2024, pp. 133--148.

\bibitem{beullens2003guided}
P.~Beullens, L.~Muyldermans, D.~Cattrysse, and D.~Van~Oudheusden, ``A guided
  local search heuristic for the capacitated arc routing problem,''
  \emph{European Journal of Operational Research}, vol. 147, no.~3, pp.
  629--643, 2003.

\bibitem{yue2023graph}
P.~Yue, S.~Liu, and Y.~Jin, ``Graph {Q}-learning assisted ant colony
  optimization for vehicle routing problems with time windows,'' in
  \emph{Proceedings of the Companion Conference on Genetic and Evolutionary
  Computation}, 2023, pp. 7--8.

\bibitem{zheng2024hybrid}
\rev{Z.~Zheng, S.~Liu, and Y.-S. Ong, ``Hybrid memetic search for electric vehicle
  routing with time windows, simultaneous pickup-delivery, and partial
  recharges,'' \emph{IEEE Transactions on Emerging Topics in Computational
  Intelligence}, vol.~9, no.~6, pp. 3773--3787, 2025.}

\bibitem{fu2025multi}
\rev{Y.~Fu, Z.~Zhang, M.~Huang, X.~Guo, and L.~Qi, ``Multi-objective integrated
  energy-efficient scheduling of distributed flexible job shop and vehicle
  routing by knowledge-and-learning-based hyper-heuristics,'' \emph{IEEE
  Transactions on Emerging Topics in Computational Intelligence}, vol.~9,
  no.~3, pp. 2137--2150, 2025.}

\bibitem{zhang2024peac}
\rev{H.~Zhang, Q.~Li, and X.~Yao, ``PEAC-HNF: A novel multi-objective evolutionary
  algorithm for split delivery vehicle routing with three-dimensional loading
  constraints,'' \emph{IEEE Transactions on Emerging Topics in Computational
  Intelligence}, vol.~9, no.~4, pp. 2830--2845, 2024.}

\bibitem{huang2025multimodal}
Y.~Huang, W.~Zhang, L.~Feng, X.~Wu, and K.~C. Tan, ``How multimodal integration
  boost the performance of LLM for optimization: Case study on capacitated
  vehicle routing problems,'' in \emph{2025 IEEE Symposium for
  Multidisciplinary Computational Intelligence Incubators (MCII)}.\hskip 1em
  plus 0.5em minus 0.4em\relax IEEE, 2025, pp. 1--7.

\bibitem{wang2023ant}
X.~Wang and Y.~Jin, ``An ant colony algorithm assisted by graph neural networks
  for solving vehicle routing problems,'' in \emph{Proceedings of the Companion
  Conference on Genetic and Evolutionary Computation}, 2023, pp. 5--6.


\bibitem{li2026novel}
Q.~Li, S.~Liu, J.~Zou, and K.~Tang, ``A novel dual-stage algorithm for
  capacitated arc routing problems with time-dependent service costs,''
  \emph{Memetic Computing}, vol.~18, no.~3, p.~33, 2026.

\bibitem{ahabchane2020mixed}
C.~Ahabchane, A.~Langevin, and M.~Tr{\'e}panier, ``The mixed capacitated
  general routing problem with time-dependent demands,'' \emph{Networks},
  vol.~76, no.~4, pp. 467--484, 2020.

\bibitem{vidal2021arc}
T.~Vidal, R.~Martinelli, T.~A. Pham, and M.~H. H{\`a}, ``Arc routing with
  time-dependent travel times and paths,'' \emph{Transportation Science},
  vol.~55, no.~3, pp. 706--724, 2021.

\bibitem{qi2015decomposition}
Y.~Qi, Z.~Hou, H.~Li, J.~Huang, and X.~Li, ``A decomposition based memetic
  algorithm for multi-objective vehicle routing problem with time windows,''
  \emph{Computers \& Operations Research}, vol.~62, pp. 61--77, 2015.

\bibitem{dong2022cell}
J.~Dong, B.~Hou, L.~Feng, H.~Tang, K.~C. Tan, and Y.-S. Ong, ``A cell-based
  fast memetic algorithm for automated convolutional neural architecture
  design,'' \emph{IEEE Transactions on Neural Networks and Learning Systems},
  vol.~34, no.~11, pp. 9040--9053, 2022.

\bibitem{boutaib2025cross}
S.~Boutaib, M.~Elarbi, S.~Bechikh, C.~A.~C. Coello, and L.~B. Said,
  ``Cross-project code smell detection as a dynamic optimization problem: An
  evolutionary memetic approach,'' in \emph{2025 IEEE Congress on Evolutionary
  Computation (CEC)}.\hskip 1em plus 0.5em minus 0.4em\relax IEEE, 2025, pp.
  1--9.

\bibitem{beke2023routing}
L.~Beke, L.~Uribe, A.~Lara, C.~A.~C. Coello, M.~Weiszer, E.~K. Burke, and
  J.~Chen, ``Routing and scheduling in multigraphs with time constraints—{A}
  memetic approach for airport ground movement,'' \emph{IEEE Transactions on
  Evolutionary Computation}, vol.~28, no.~2, pp. 474--488, 2023.

\bibitem{wen2024transfer}
X.~Wen, G.~Wu, J.~Liu, and Y.-S. Ong, ``Transfer optimization for heterogeneous
  drone delivery and pickup problem,'' \emph{IEEE Transactions on Emerging
  Topics in Computational Intelligence}, vol.~9, no.~1, pp. 347--364, 2024.

\bibitem{liu2023good}
S.~Liu, Y.~Zhang, K.~Tang, and X.~Yao, ``How good is neural combinatorial
  optimization? {A} systematic evaluation on the traveling salesman problem,''
  \emph{IEEE Computational Intelligence Magazine}, vol.~18, no.~3, pp. 14--28,
  2023.

\bibitem{shang2018memetic}
R.~Shang, B.~Du, K.~Dai, L.~Jiao, and Y.~Xue, ``Memetic algorithm based on
  extension step and statistical filtering for large-scale capacitated arc
  routing problems,'' \emph{Natural Computing}, vol.~17, pp. 375--391, 2018.

\bibitem{2021Memetic}
R.~Li, X.~Zhao, X.~Zuo, J.~Yuan, and X.~Yao, ``Memetic algorithm with
  non-smooth penalty for capacitated arc routing problem,''
  \emph{Knowledge-Based Systems}, vol. 220, p. 106957, 2021.


\bibitem{lan2021region}
W.~Lan, Z.~Ye, P.~Ruan, J.~Liu, P.~Yang, and X.~Yao, ``Region-focused memetic
  algorithms with smart initialization for real-world large-scale waste
  collection problems,'' \emph{IEEE Transactions on Evolutionary Computation},
  vol.~26, no.~4, pp. 704--718, 2021.

\bibitem{dijkstra1959note}
E.~DIJKSTRA, ``A note on two problems in connexion with graphs,''
  \emph{Numerische Mathematik}, vol.~1, pp. 269--271, 1959.

\bibitem{OVERHOLT1967FIBONACCI}
K.~OVERHOLT, ``Fibonacci search - minx and golden section search,''
  \emph{COMPUTER JOURNAL}, vol.~9, no.~4, pp. 414--\&, 1967.

\bibitem{tang2016negatively}
K.~Tang, P.~Yang, and X.~Yao, ``Negatively correlated search,'' \emph{IEEE
  Journal on Selected Areas in Communications}, vol.~34, no.~3, pp. 542--550,
  2016.

\bibitem{runarsson2000stochastic}
T.~P. Runarsson and X.~Yao, ``Stochastic ranking for constrained evolutionary
  optimization,'' \emph{IEEE Transactions on evolutionary computation}, vol.~4,
  no.~3, pp. 284--294, 2000.

\bibitem{golden1983computational}
B.~L. Golden, J.~S. DeArmon, and E.~K. Baker, ``Computational experiments with
  algorithms for a class of routing problems,'' \emph{Computers \& Operations
  Research}, vol.~10, no.~1, pp. 47--59, 1983.

\bibitem{eglese1994routeing}
R.~W. Eglese, ``Routeing winter gritting vehicles,'' \emph{Discrete applied
  mathematics}, vol.~48, no.~3, pp. 231--244, 1994.

\bibitem{solomon1987algorithms}
M.~M. Solomon, ``Algorithms for the vehicle routing and scheduling problems
  with time window constraints,'' \emph{Operations research}, vol.~35, no.~2,
  pp. 254--265, 1987.

\bibitem{wilcoxon1992individual}
F.~Wilcoxon, ``Individual comparisons by ranking methods,'' in
  \emph{Breakthroughs in statistics: Methodology and distribution}.\hskip 1em
  plus 0.5em minus 0.4em\relax Springer, 1992, pp. 196--202.

\bibitem{liu2021memetic}
S.~Liu, K.~Tang, and X.~Yao, ``Memetic search for vehicle routing with
  simultaneous pickup-delivery and time windows,'' \emph{Swarm and Evolutionary
  Computation}, vol.~66, p. 100927, 2021.

\bibitem{guo2025learning}
\rev{T.~Guo, Y.~Mei, M.~Zhang, H.~Zhao, K.~Cai, and W.~Du, ``Learning-aided
  neighborhood search for vehicle routing problems,'' \emph{IEEE Transactions
  on Pattern Analysis and Machine Intelligence}, vol.~47, no.~7, pp.
  5930--5944, 2025.}

\bibitem{wang2024deep}
\rev{C.~Wang, M.~Cao, H.~Jiang, X.~Xiang, and X.~Zhang, ``A deep reinforcement
  learning-based adaptive large neighborhood search for capacitated electric
  vehicle routing problems,'' \emph{IEEE Transactions on Emerging Topics in
  Computational Intelligence}, vol.~9, no.~1, pp. 131--144, 2024.}

\bibitem{liu2024large}
S.~Liu, C.~Chen, X.~Qu, K.~Tang, and Y.-S. Ong, ``Large language models as
  evolutionary optimizers,'' in \emph{2024 IEEE Congress on Evolutionary
  Computation (CEC)}.\hskip 1em plus 0.5em minus 0.4em\relax IEEE, 2024, pp.
  1--8.

\bibitem{hou2017evolutionary}
Y.~Hou, Y.-S. Ong, L.~Feng, and J.~M. Zurada, ``An evolutionary transfer
reinforcement learning framework for multiagent systems,'' \emph{IEEE
Transactions on Evolutionary Computation}, vol.~21, no.~4, pp. 601--615,
2017.

\bibitem{hou2021advances}
Y.~Hou, X.~Yu, Y.~Zeng, Z.~Wei, H.~Zhang, H.~Ge, and Q.~Zhang, ``Advances in
memetic automaton: Toward human-like autonomous agents in complex multi-agent
learning problems,'' \emph{IEEE Computational Intelligence Magazine},
vol.~16, no.~4, pp. 54--69, 2021.

\bibitem{meng2026uniteformer}
D.~Meng, Z.~Cao, J.~Gao, Y.~Wu, and Y.~Hou, ``Uniteformer: Unifying node and
edge modalities in transformers for vehicle routing problems,''
\emph{Advances in Neural Information Processing Systems}, vol.~38, pp.
93\,830--93\,854, 2026.
\end{thebibliography}

\end{document}